\documentclass[10pt,twocolumn,letterpaper]{article}

\usepackage{cvpr}               %

\usepackage[T1]{fontenc}
\usepackage{cuted}      %
\usepackage{currfile}   %
\usepackage{caption}    %
\usepackage{multirow}
\usepackage{makecell}
\usepackage{colortbl}
\usepackage{tikz}
\usetikzlibrary{positioning,arrows.meta,fit,backgrounds,calc,shapes.geometric}
\usepackage{microtype}

\DeclareMathOperator*{\argmin}{arg\,min}

\makeatletter
\renewcommand\paragraph{\@startsection{paragraph}{4}{\z@}{\baselineskip}{-1em}{\normalfont\normalsize\bfseries}}
\makeatother

\definecolor{cvprblue}{rgb}{0.21,0.49,0.74}
\usepackage[pagebackref,breaklinks,colorlinks,allcolors=cvprblue]{hyperref}

\def\confName{CVPR}
\def\confYear{2027}

\title{JEPA4R: Embodied JEPA World Models with 4D Point Planning}

\author{
Jiarui Dong$^{1*}$ \quad
Zhiheng Liang$^{1*}$ \quad
Yanjun Li$^{1*}$ \quad
Zeyu Zhang$^{2*\dag}$ \\
Heejeong Nam$^{3}$ \quad
Haoyu Chen$^{4}$ \quad
Hao Tang$^{1\ddag}$ \\
$^{1}$Peking University \quad
$^{2}$UC Berkeley BAIR\quad
$^{3}$Brown University \quad
$^{4}$University of Oulu \\
\small $^*$Equal contribution.
$^\dag$Project lead.
$^\ddag$Corresponding author: bjdxtanghao@gmail.com.
}

\begin{document}
\crefname{subtable}{Tab.}{Tabs.}\Crefname{subtable}{Table}{Tables}
\maketitle
\begin{strip}
\centering
\noindent\begin{tikzpicture}[x=1pt,y=1pt,font=\fontsize{6.67}{8}\selectfont]
\useasboundingbox (0,0) rectangle (149,120);
\node[anchor=south west,inner sep=0] at (0,0) {\includegraphics[height=120pt]{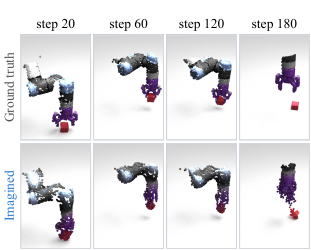}};
\node[anchor=north west,inner sep=0] at (0,120) {\textbf{(a)}~Imagined rollouts stay within 1.1\,cm over 200 steps};
\end{tikzpicture}\hfill
\begin{tikzpicture}[x=1pt,y=1pt,font=\fontsize{6.67}{8}\selectfont]
\useasboundingbox (0,0) rectangle (203.6,120);
\node[anchor=north,inner sep=0] at (101.8,120) {\textbf{(b)}~JEPA4R outperforms the best 2D JEPA in policy success};
\node[anchor=south west,inner sep=0] at (0,0) {\includegraphics[width=203.6pt,height=113pt]{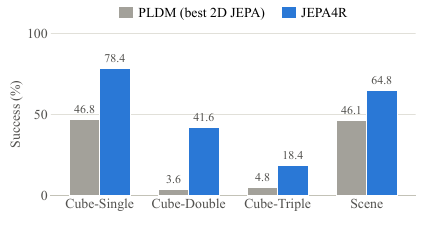}};
\end{tikzpicture}\hfill
\begin{tikzpicture}[x=1pt,y=1pt,font=\fontsize{6.67}{8}\selectfont]
\useasboundingbox (0,0) rectangle (125.8,120);
\node[anchor=north west,inner sep=0] at (0,120) {\textbf{(c)}~Twin-trained, zero-shot on the real arm};
\node[anchor=south west,inner sep=0] at (5.0,51.2) {\includegraphics[width=37.6pt,height=48.4pt]{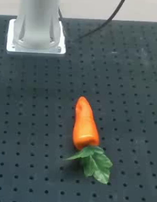}};
\node[anchor=south west,inner sep=0] at (44.1,51.2) {\includegraphics[width=37.6pt,height=48.4pt]{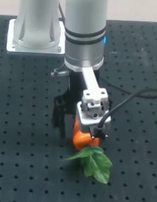}};
\node[anchor=south west,inner sep=0] at (83.2,51.2) {\includegraphics[width=37.6pt,height=48.4pt]{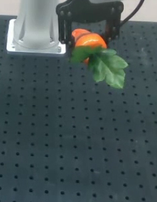}};
\node[anchor=south west,inner sep=0] at (5.0,1.2) {\includegraphics[width=37.6pt,height=48.4pt]{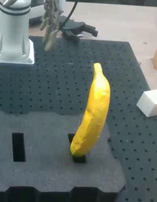}};
\node[anchor=south west,inner sep=0] at (44.1,1.2) {\includegraphics[width=37.6pt,height=48.4pt]{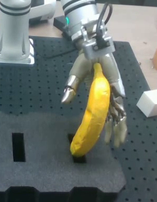}};
\node[anchor=south west,inner sep=0] at (83.2,1.2) {\includegraphics[width=37.6pt,height=48.4pt]{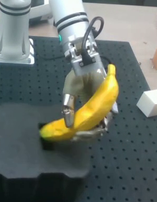}};
\end{tikzpicture}
\captionsetup{hypcap=false}
\captionof{figure}{\textbf{JEPA4R, a JEPA world model on single-camera point clouds, outperforms 2D JEPAs in policy learning, decodes its imagined rollouts into 4D point clouds, and runs zero-shot on the real arm after training only in a digital twin.} (a)~Ground truth \vs point clouds decoded from latents predicted from three context frames and actions (OGBench Cube; gripper purple, cube red). (b)~JEPA4R \vs\ the best 2D JEPA (PLDM) on Cube-Single, Cube-Double, Cube-Triple and Scene. Numbers from \cref{tab:ogbench_il}. (c)~Real-robot deployment on an RM75-B arm, with a parallel gripper (top) and a dexterous hand (bottom).}
\captionsetup{hypcap=true}
\vspace{-.7em}
\label{fig:teaser}
\end{strip}
\begin{abstract}
Joint-embedding predictive architecture (JEPA) world models predict future embeddings instead of pixels and plan in latent space.
Two challenges remain: 2D images leave the metric geometry that manipulation acts on implicit and entangled with appearance, and what these models imagine stays hidden, as they are mostly judged in simulation by short-horizon latent planning.
We present \textbf{JEPA4R}, a JEPA world model that observes single-depth-camera point clouds, predicts and plans in their latent space, and acts through a policy head on its frozen latents; we call this 4D point planning.
A decoder on the frozen world model shows what it imagines as 4D point clouds, with a mean point error within 1.1\,cm over 200 steps.
Under the policy-learning protocols of OGBench and CortexBench, JEPA4R leads pixel JEPA models on OGBench Cube-Single, Cube-Double, Cube-Triple and Scene, and ranks first on MetaWorld and Adroit Relocate with \textbf{98.4\%} and \textbf{88.0\%}, above the runner-up, R3M, at 96.0\% and 65.3\%.
Trained only in a digital twin, JEPA4R and its policy heads run zero-shot on the real arm with both a gripper and a dexterous hand.
\href{https://github.com/AIGeeksGroup/JEPA4R}{Code} \& 
\href{https://aigeeksgroup.github.io/JEPA4R}{Website}.
\end{abstract}

\section{Introduction}\label{sec:intro}

World models built on joint-embedding predictive architectures (JEPAs) predict how a scene's latent changes under an action and plan toward the latent of a goal~\cite{assran2025vjepa2,zhou2025dinowm,sobal2025pldm}.
However, current JEPA world models have two significant challenges in robot manipulation, which hinges on metric 3D geometry:

First, most JEPA world models read 2D input. Point-cloud JEPAs mostly pretrain encoders without actions~\cite{saito2025pointjepa,hu2024jepa3d,lee2026pointtube,zhu2026adljepa} or forecast LiDAR occupancy~\cite{zhu2026adlistjepa}, so metric geometry rarely reaches the model that predicts and acts.
An image entangles metric geometry with appearance and perspective, so a pixel model must infer depth, whereas a point cloud from the same camera records visible surfaces at their metric positions (\cref{fig:obs}).
Even web-scale 2D foundation features show limited 3D awareness~\cite{elbanani2024probe3d}.
LeWM~\cite{maes2026lewm}, trained from pixels without pretraining, has its lowest success on OGBench Cube~\cite{park2025ogbench}, where it trails DINO-WM~\cite{zhou2025dinowm} on frozen DINOv2 features (74\% \vs 86\%)~\cite{maes2026lewm}.
Its authors relate this gap partly to the 3D nature of the scene~\cite{maes2026lewm}.\looseness=1

\looseness=1 Second, what these models imagine stays hidden. JEPA latents are trained to be predictable, not reconstructive. Existing tools only decode them to images~\cite{zhou2025dinowm,maes2026lewm} or audit them indirectly~\cite{maes2026lewm,walker2026jacobian}, so the imagined 3D scene stays out of sight, which hampers debugging and real-robot safety checks. Scores do not reveal it either.
JEPA world models are mostly scored in simulation by short-horizon goal reaching~\cite{zhou2025dinowm,sobal2025pldm,maes2026lewm}, where a sampling-based planner such as the cross-entropy method~\cite{deboer2005cem} searches a few action blocks for the predicted latent closest to a goal latent.
Such search tests the predictor over short, in-distribution horizons, not how well the latents serve the policies robots learn. For example, LeWM latents that plan well transfer poorly to policy learning~\cite{liu2026tcsigreg}, and the model is evaluated only in simulation~\cite{maes2026lewm}.
Frozen-encoder imitation on established suites such as CortexBench~\cite{majumdar2023vc1} asks instead how well a representation lets a policy learn a skill from a few demonstrations, which matters when robot data are scarce. However, JEPA world models rarely face it.

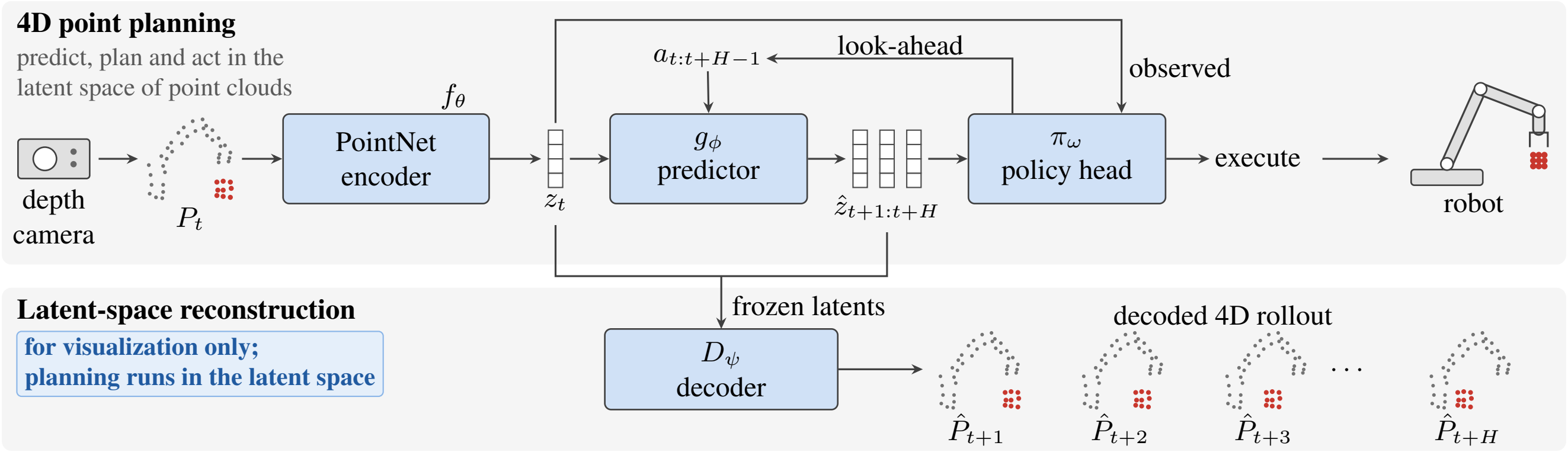
\begin{figure*}[t]
\centering
\definecolor{jmnew}{HTML}{2A78D6}%
\definecolor{jminh}{RGB}{140,140,140}%
\definecolor{jmobj}{HTML}{C8372D}%
\newcommand{\jmcloud}[4]{%
  \begin{scope}[shift={(#1,#2)}, scale=#4]
    \foreach \px/\py in {0.09/0.05,0.11/0.15,0.10/0.26,0.08/0.37,0.24/0.07,0.24/0.15,0.26/0.29,0.24/0.36,%
        0.16/0.01,0.23/0.00,0.22/0.47,0.29/0.56,0.33/0.62,0.41/0.73,0.47/0.79,0.57/0.86,0.28/0.44,0.34/0.52,%
        0.45/0.61,0.51/0.69,0.60/0.75,0.68/0.88,0.72/0.81,0.79/0.76,0.86/0.67,0.93/0.59,0.69/0.78,0.73/0.70,%
        0.79/0.65,0.88/0.58,0.87/0.49,0.86/0.45,0.86/0.40,0.99/0.51,0.98/0.45,0.97/0.41}
      {\fill[black!55] (\px,\py) circle (0.6pt);}
    \foreach \px/\py in {0.84/0.03,0.85/0.09,0.84/0.18,0.91/0.02,0.91/0.09,0.91/0.19,0.99/0.01,1.00/0.11,0.99/0.18}
      {\fill[jmobj] ({\px+#3},\py) circle (0.8pt);}
  \end{scope}}
\newcommand{\jmlatent}[2]{%
  \foreach \k in {0,1,2,3}
    {\draw[black!70, fill=white, line width=0.4pt] ({#1-0.08},{#2-0.32+0.16*\k}) rectangle ++(0.16,0.16);}}
\begin{tikzpicture}[
    font=\small,
    >={Stealth[length=3.6pt,width=3.2pt]},
    box/.style={draw=black!75, line width=0.5pt, rounded corners=2pt, align=center,
                minimum height=10mm, inner sep=2pt},
    mod/.style={box, fill=jmnew!22},
    arr/.style={->, line width=0.6pt, draw=black!75},
    line/.style={line width=0.6pt, draw=black!75},
    lbl/.style={inner sep=1pt, align=center},
    band/.style={fill=black!4, rounded corners=3pt},
    ttl/.style={inner sep=0pt, anchor=north west, align=left, font=\small\bfseries},
    sub/.style={inner sep=0pt, anchor=north west, align=left, font=\footnotesize, text=black!65},
    hl/.style={anchor=north west, align=left, font=\footnotesize\bfseries, text=jmnew!75!black, fill=jmnew!12, draw=jmnew!55, line width=0.4pt, rounded corners=1.5pt, inner xsep=2.5pt, inner ysep=2pt},
  ]
  \fill[band] (-0.18,-1.18) rectangle (17.28,1.76);
  \fill[band] (-0.18,-1.44) rectangle (17.28,-3.26);
  \node[ttl] at (-0.02,1.64) {4D point planning};
  \node[sub] at (-0.02,1.24) {predict, plan and act in the\\latent space of point clouds};
  \draw[black!75, fill=black!12, rounded corners=1pt, line width=0.5pt] (0,-0.22) rectangle (0.8,0.22);
  \draw[black!75, fill=white, line width=0.5pt] (0.30,0) circle (0.13);
  \fill[black!60] (0.62,0.08) circle (0.035);
  \fill[black!60] (0.62,-0.06) circle (0.035);
  \node[lbl, below] at (0.40,-0.30) {depth\\camera};
  \draw[arr] (0.90,0) -- (1.34,0);
  \jmcloud{1.38}{-0.44}{0}{1}
  \node[lbl, below] at (1.92,-0.50) {$P_t$};
  \draw[arr] (2.50,0) -- (2.95,0);
  \node[mod, minimum width=2.3cm, anchor=west] (enc) at (2.95,0) {PointNet\\encoder};
  \node[lbl, above] at ([xshift=-0.40cm]enc.north east) {$f_\theta$};
  \draw[arr] (enc.east) -- (5.84,0);
  \jmlatent{6.00}{0}
  \node[lbl, below] at (6.00,-0.36) {$z_t$};
  \draw[arr] (6.16,0) -- (6.60,0);
  \node[mod, minimum width=2.2cm, anchor=west] (pred) at (6.60,0) {$g_\phi$\\predictor};
  \node[lbl] (act) at (7.70,1.12) {$a_{t:t+H-1}$};
  \draw[arr] (act.south) -- (pred.north);
  \draw[arr] (pred.east) -- (9.24,0);
  \jmlatent{9.40}{0}
  \jmlatent{9.70}{0}
  \jmlatent{10.00}{0}
  \node[lbl, below] at (9.70,-0.36) {$\hat z_{t+1:t+H}$};
  \draw[arr] (10.16,0) -- (10.60,0);
  \node[mod, minimum width=2.2cm, anchor=west] (pol) at (10.60,0) {$\pi_\omega$\\policy head};
  \draw[arr] (6.00,0.40) -- (6.00,1.55) -| ([xshift=-0.50cm]pol.north east);
  \draw[arr] ([xshift=0.50cm]pol.north west) |- (act.east);
  \node[lbl, above] at (9.85,1.12) {look-ahead};
  \node[lbl, anchor=west] at (12.36,1.02) {observed};
  \draw[arr] (pol.east) -- (13.30,0);
  \node[lbl, anchor=west] at (13.32,0) {execute};
  \draw[arr] (14.56,0) -- (15.30,0);
  \begin{scope}[shift={(15.95,-0.12)}, scale=1.25]
    \draw[black!75, fill=black!12, rounded corners=1pt, line width=0.5pt] (-0.32,-0.14) rectangle (0.32,0);
    \foreach \a/\b in {{(0,0.05)}/{(0.30,0.72)}, {(0.30,0.72)}/{(0.82,0.52)}, {(0.82,0.52)}/{(0.82,0.36)}}
      {\draw[black!75, line width=4.2pt, line cap=round] \a -- \b;
       \draw[black!12, line width=3.0pt, line cap=round] \a -- \b;}
    \foreach \p in {(0,0.05), (0.30,0.72), (0.82,0.52)}
      {\draw[black!75, fill=white, line width=0.5pt] \p circle (0.055);}
    \draw[black!75, line width=0.6pt] (0.74,0.20) -- (0.74,0.33) -- (0.90,0.33) -- (0.90,0.20);
    \foreach \px/\py in {0.77/0.03,0.82/0.03,0.87/0.03,0.77/0.08,0.82/0.08,0.87/0.08,0.77/0.13,0.82/0.13,0.87/0.13}
      {\fill[jmobj] (\px,\py) circle (0.8pt);}
  \end{scope}
  \node[lbl, below] at (16.25,-0.34) {robot};
  \node[ttl] at (-0.02,-1.56) {Latent-space reconstruction};
  \node[hl] at (-0.02,-1.92) {for visualization only;\\planning runs in the latent space};
  \draw[line] (6.00,-0.74) -- (6.00,-1.31) -- (9.70,-1.31) -- (9.70,-0.82);
  \draw[arr] (7.85,-1.31) -- (7.85,-1.90);
  \node[lbl, anchor=west] at (7.93,-1.64) {frozen latents};
  \node[mod, minimum width=2.6cm, minimum height=9mm] (dec) at (7.85,-2.35) {$D_\psi$\\decoder};
  \draw[arr] (dec.east) -- (10.10,-2.35);
  \node[lbl, anchor=south] at (13.45,-1.90) {decoded 4D rollout};
  \jmcloud{10.22}{-2.78}{0.00}{0.95}
  \jmcloud{11.82}{-2.78}{-0.15}{0.95}
  \jmcloud{13.42}{-2.78}{-0.30}{0.95}
  \node at (14.86,-2.36) {$\cdots$};
  \jmcloud{15.70}{-2.78}{-0.45}{0.95}
  \node[lbl, below] at (10.70,-2.80) {$\hat P_{t+1}$};
  \node[lbl, below] at (12.30,-2.80) {$\hat P_{t+2}$};
  \node[lbl, below] at (13.90,-2.80) {$\hat P_{t+3}$};
  \node[lbl, below] at (16.18,-2.80) {$\hat P_{t+H}$};
\end{tikzpicture}
\caption{\textbf{JEPA4R predicts, plans and acts in the latent space of single-depth-camera point clouds and decodes what it imagines into 4D point clouds.} Top (4D point planning): the encoder $f_\theta$ maps $P_t$ to $z_t$, the predictor $g_\phi$ rolls the latent forward under actions, and the policy head $\pi_\omega$ acts on $z_t$ or, to look ahead, on imagined latents. Bottom (latent-space reconstruction): the decoder $D_\psi$, attached to the frozen world model, turns every imagined rollout into 4D point clouds (\cref{eq:decoder}).}
\label{fig:method}
\end{figure*}

\looseness=1 To address these two challenges, we propose JEPA4R, a JEPA world model designed to see metric geometry and to show what it imagines, so that a robot's predictions can be checked before it acts. It reads a single-camera point cloud through a parameter-matched encoder (\cref{fig:method}).
Every point cloud is paired with an image from the same eye-to-hand camera, so the 2D JEPAs are trained or evaluated on the same trajectories and viewpoint.
To evaluate how JEPA4R performs in robot policy learning, we train a policy head on the frozen latents under established protocols. Specifically, we use goal-conditioned policy learning on OGBench~\cite{park2025ogbench} and imitation on CortexBench, where JEPA4R is ranked among published representations and 2D JEPAs evaluated on the same data.
JEPA4R predicts, plans and acts in the 4D point latent space of its encoder, and we call this 4D point planning. An imagined rollout is the latent sequence that the predictor generates autoregressively from context frames and actions. A decoder attached to the frozen world model shows this imagination by turning every imagined rollout into a decoded 4D rollout (\cref{fig:teaser}a).
All world models train in simulation. The RM75-B twin, a digital twin of the RealMan RM75-B arm, lets JEPA4R's twin-trained world models and policy heads run zero-shot on the real arm.
Notably, we evaluate policy learning on manipulation suites with colored point clouds from a pipeline that also runs on the real arm, while previous work~\cite{oberweger2026pointlewm} evaluates planning alone with xyz-only raycast point clouds and finds the two modalities on par.

JEPA4R leads the 2D JEPAs on OGBench Cube and Scene and ranks first on CortexBench, above every published representation (\cref{tab:ogbench_il,tab:metaworld}, \cref{fig:ogbench_plots}). Its decoder recovers the observed scene from encoded and predicted latents alike, so even long imagined rollouts stay close to the observed point clouds (\cref{tab:recon}, \cref{fig:recon}). On the real arm, JEPA4R's twin-trained policies run with both a parallel gripper and a dexterous hand.

In general, our paper can be summarized in the following three points:

\begin{itemize}
\item 4D point planning: we propose JEPA4R, which predicts, plans and acts in the latent space of realistic single-depth-camera point clouds, to bring the metric 3D geometry that 2D inputs leave implicit into JEPA world models.
\item Latent-space reconstruction: we propose a decoder attached to the frozen world model that turns every imagined rollout into 4D point clouds, to show what JEPA world models imagine, which their latents keep hidden. The decoded imagination keeps a mean point error within 1.1\,cm over 200 environment steps, so it can be inspected for debugging and real-robot safety checks.
\item Experiments on benchmarks and in the real world: we learn policies on OGBench and CortexBench in a paired 2D/3D evaluation, where JEPA4R's point-cloud features rank first on CortexBench, above every published representation and 2D JEPA. JEPA4R transfers zero-shot to the real arm with both a gripper and a dexterous hand, after training only in a digital twin.
\end{itemize}
\section{Related work}
\label{sec:related}

\paragraph{JEPA and latent world models.}
\looseness=1 JEPA world models extend joint-embedding prediction~\cite{assran2023ijepa,bardes2024vjepa,garrido2024iwm,mo2024cjepa,murlabadia2026vjepa21,balestriero2025lejepa,wu2026visreg} to action-conditioned dynamics and regress latents, unlike world models that generate observations~\cite{hafner2023dreamerv3,bar2025nwm} or predict rewards~\cite{hansen2024tdmpc2}.
DINO-WM~\cite{zhou2025dinowm} and V-JEPA~2-AC~\cite{assran2025vjepa2} learn predictors on frozen foundation features, whereas PLDM~\cite{sobal2025pldm} and LeWM~\cite{maes2026lewm} train encoder and predictor end to end.
Follow-up work analyzes planning success~\cite{terver2026jepawms,alrasheed2026planninglimits}, generalization beyond expert rollouts~\cite{nam2026anygoal}, counterfactual action evaluation~\cite{subramanian2026counterfactual} and behavioral monitoring~\cite{walker2026jacobian}, and adds objectives for action sensitivity, physical-state grounding~\cite{zhang2026deltajepa,zeng2026phylatent,yan2026psgjepa}, object-level masking~\cite{nam2026cjepa} and policy transfer~\cite{liu2026tcsigreg}, but all these models observe images, whose features show limited 3D awareness~\cite{elbanani2024probe3d}.
JEPA world models are mostly scored by planning success~\cite{zhou2025dinowm,sobal2025pldm,maes2026lewm,maes2026stableworldmodel}, and PSG-JEPA~\cite{yan2026psgjepa} adds policy learning, whereas pretrained robot representations~\cite{nair2022r3m,xiao2022mvp,radosavovic2022mvp,ma2023vip,karamcheti2023voltron,shang2024theia,yang2026stp} are compared by frozen-feature imitation, chiefly in CortexBench~\cite{majumdar2023vc1} on MetaWorld~\cite{yu2019metaworld} and Adroit~\cite{rajeswaran2018dapg}, with further studies varying learner, data and test conditions~\cite{parisi2022pvr,dittadi2022ood,hansen2023lfs,hu2023notallequal,dasari2023unbiased,silwal2024pvr,burns2024robustpvr,schneider2024pvrmbrl,dong2026structure}.
We learn policies on OGBench~\cite{park2025ogbench} and CortexBench with paired 2D/3D data and compare them with published representations and 2D JEPAs.

\paragraph{3D world models.}
Most point-cloud JEPAs pretrain encoders without actions~\cite{saito2025pointjepa,hu2024jepa3d,lee2026pointtube,zhu2026adljepa} or forecast LiDAR occupancy~\cite{zhu2026adlistjepa} instead of driving a planner.
Other 3D world models forecast explicit particles~\cite{li2019dpinet,whitney2024vpd}, points~\cite{huang2025particleformer,peri2026pointworldmodels,huang2026pointworld,zhang2024copilot4d}, Gaussians~\cite{lu2025gwm}, depth-augmented video~\cite{zhen2025tesseract}, structured latents~\cite{li2026structured4d} or occupancy~\cite{zheng2024occworld}, and 3DPWM~\cite{peri2026pointworldmodels} and PointWorld~\cite{huang2026pointworld} plan with such forecasts on real arms rather than in a latent space. Such explicit forecasts make every prediction visible at the cost of generating each element of the next scene, whereas JEPA world models predict in a latent space and have so far left the imagined scene unseen.
On the policy side, point-cloud encoders range from PointNet~\cite{qi2017pointnet,qi2017pointnetpp} to Point Transformer V3~\cite{wu2024ptv3} and Sonata~\cite{wu2025sonata}, 3D policies condition diffusion heads on such encoders~\cite{ze2024dp3,ze2024idp3,ke2024diffuseractor}, and controlled studies favor point clouds and 3D-aware encoders over 2D inputs~\cite{zhu2024pointcloudmatters,gyenes2024pprl,hong2026r3d,zhu2025spa,qian2025mvp3d,liang2026afro}, but such policies lack a world model.
Decoders on frozen features map self-supervised representations~\cite{bordes2022rcdm,zheng2026rae} and the predicted latents of JEPA world models~\cite{zhou2025dinowm,maes2026lewm} to images, and recent work lifts video world-model rollouts into 3D through generated pixels~\cite{kim2026rogsw4rld,wang2026robophys3d}.
Feed-forward networks regress dense point maps from images~\cite{wang2024dust3r,leroy2024mast3r,wang2025vggt}, PAGE-4D~\cite{zhou2026page4d} extends VGGT to dynamic scenes, and point-set decoders generate or complete clouds~\cite{fan2017psgn,achlioptas2018pcae,yang2018foldingnet,yuan2018pcn,yu2021pointr}.
Closest to ours, Oberweger~\etal~\cite{oberweger2026pointlewm} find LeWM on idealized raycast point clouds statistically on par with its pixel version on its own benchmarks.
JEPA4R instead reads colored point clouds from one fixed depth camera. Its policy head acts on the frozen latents of a world model that can also imagine the scene, and a decoder adapted from PAGE-4D turns these latents directly into point clouds.

\paragraph{Sim-to-real with 3D observations.}
Depth and point-cloud observations narrow the gap between simulation and reality: policies trained on them in simulation transfer to real hands and arms~\cite{qin2022dexpoint,chen2023visualdexterity,liu2025cdm}, point-cloud policies run on real arms and humanoids~\cite{ze2024dp3,ze2024idp3}, and point-cloud models resist changes in appearance and viewpoint~\cite{peri2024pcwm,ze2024dp3}.
Simulators are also matched to the deployment scene. RialTo~\cite{torne2024rialto} scans a digital twin and robustifies point-cloud policies in it, ACDC~\cite{dai2024acdc} generates digital cousins from one image for zero-shot transfer, and Gaussian splatting renders photorealistic RGB for sim-to-real~\cite{qureshi2024splatsim,li2024robogsim}.
World models are either learned directly on physical robots~\cite{wu2022daydreamer} or trained in simulation and transferred to legged robots~\cite{lai2024wmp,li2025rwm} and arms~\cite{peri2026pointworldmodels,wang2026sim2realwam}.
Among JEPA world models, V-JEPA~2-AC~\cite{assran2025vjepa2} plans on real arms with a predictor trained on real robot video, whereas DINO-WM~\cite{zhou2025dinowm} and LeWM~\cite{maes2026lewm} are evaluated in simulation. JEPA4R is a point-cloud JEPA world model trained in a twin of its deployment arm.

\section{Preliminaries}
\label{sec:prelim}

\looseness=1 Following LeWM~\cite{maes2026lewm}, we train world models on offline trajectories of observations $o_t$ and actions, cut into clips of four frames spaced five environment steps apart. The action block $a_t$ concatenates the five actions between consecutive frames. An encoder with a projector, $f_\theta$, maps each observation to a latent $z_t=f_\theta(o_t)\in\mathbb{R}^{192}$. The pixel encoder is a ViT-tiny~\cite{dosovitskiy2021vit} trained from scratch on $224\times224$ images. The predictor $g_\phi$, a six-block transformer conditioned on action embeddings through AdaLN-zero~\cite{peebles2023dit}, attends causally over at most three past latents; within a clip, $\hat z_{t+1}=g_\phi(z_{1:t},a_{1:t})$. With teacher forcing on a clip $o_{1:4}$, the model minimizes
\begin{equation}
  \mathcal{L}(\theta,\phi)=\frac{1}{3d}\sum_{t=1}^{3}\big\lVert\hat z_{t+1}-z_{t+1}\big\rVert_2^2+\lambda\,\mathrm{SIGReg}(Z),
  \label{eq:objective}
\end{equation}
\looseness=0 with $d=192$, $Z$ the batch of latents and $\lambda=0.09$ in our runs, without a stop-gradient or an EMA target encoder. The first term makes latents predictable from actions, and SIGReg~\cite{balestriero2025lejepa}, which compares 1,024 random one-dimensional projections of each time step's latents with a standard Gaussian, prevents collapse.

\looseness=1 The model acts by planning: the cross-entropy method (CEM)~\cite{deboer2005cem} searches for the sequence of $H=5$ action blocks whose last predicted latent lies closest to the latent of a goal observation. We follow this planning setting in the supplementary material (\cref{sec:supp-planning}) and evaluate every world model through a policy head on its frozen latents (\cref{sec:method-policy}).
\section{Method: JEPA4R}
\label{sec:method}

Manipulation acts on metric geometry, and a robot should see what its world model expects before it moves. JEPA4R therefore observes the scene as a point cloud from one depth camera, predicts and plans in the latent space of these point clouds, acts through a policy head on its frozen latents, and decodes what it imagines into 4D point clouds (\cref{fig:method}).

\subsection{Point clouds from one depth camera}
\label{sec:method-pcd}

\begin{figure}[t]
\centering
\definecolor{c2d}{HTML}{5F5C58}\definecolor{c3d}{HTML}{2A78D6}%
\definecolor{axr}{HTML}{D1495B}\definecolor{axg}{HTML}{2E9B6A}\definecolor{axb}{HTML}{2A78D6}%
\begin{tikzpicture}[x=1pt,y=1pt,font=\fontsize{6.67}{8}\selectfont]
\useasboundingbox (0,0) rectangle (235.0,112.0);
\begin{scope}
  \clip[rounded corners=2.5pt] (39.88,0) rectangle ++(155.23,112.00);
  \node[anchor=south west,inner sep=0] at (39.88,0) {\includegraphics[width=155.23pt,height=112.00pt]{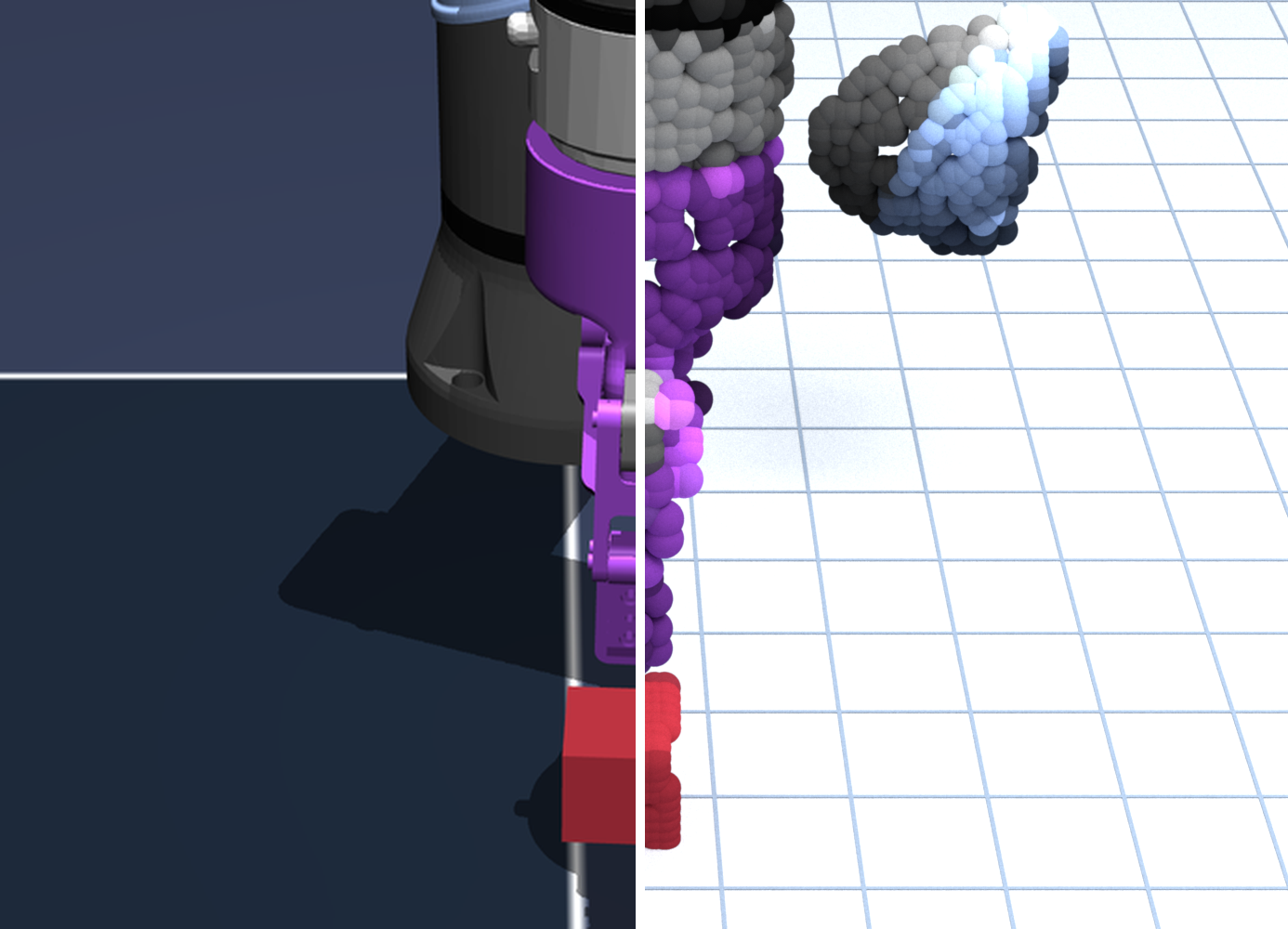}};
\end{scope}
\draw[c2d, line width=0.9pt, rounded corners=2.5pt] (116.94,112.00) -- (39.88,112.00) -- (39.88,0) -- (116.94,0);
\draw[c3d, line width=0.9pt, rounded corners=2.5pt] (116.94,112.00) -- (195.12,112.00) -- (195.12,0) -- (116.94,0);
\node[anchor=east, align=right, text=c2d] at (36.88,56.00) {\textbf{Pixels}\\$\to$ 2D JEPAs};
\node[anchor=west, align=left, text=c3d] at (198.12,56.00) {\textbf{Point cloud}\\$\to$ JEPA4R};
\draw[-{Stealth[length=3pt,width=2.6pt]}, line width=0.8pt, draw=axr] (165.38,33.08) -- (167.70,23.40) node[below, inner sep=0.6pt, font=\fontsize{6}{7}\selectfont\itshape, text=axr] {x};
\draw[-{Stealth[length=3pt,width=2.6pt]}, line width=0.8pt, draw=axg] (165.38,33.08) -- (180.78,33.08) node[right, inner sep=0.6pt, font=\fontsize{6}{7}\selectfont\itshape, text=axg] {y};
\draw[-{Stealth[length=3pt,width=2.6pt]}, line width=0.8pt, draw=axb] (165.38,33.08) -- (167.24,45.72) node[above, inner sep=0.6pt, font=\fontsize{6}{7}\selectfont\itshape, text=axb] {z};
\end{tikzpicture}
\caption{\textbf{One scene, two observations.} One OGBench Cube-Single frame seen by one camera, as pixels (left of the white line), the input of 2D JEPAs, and as a 1,200-point cloud (right), the input of JEPA4R.}
\label{fig:obs}
\end{figure}
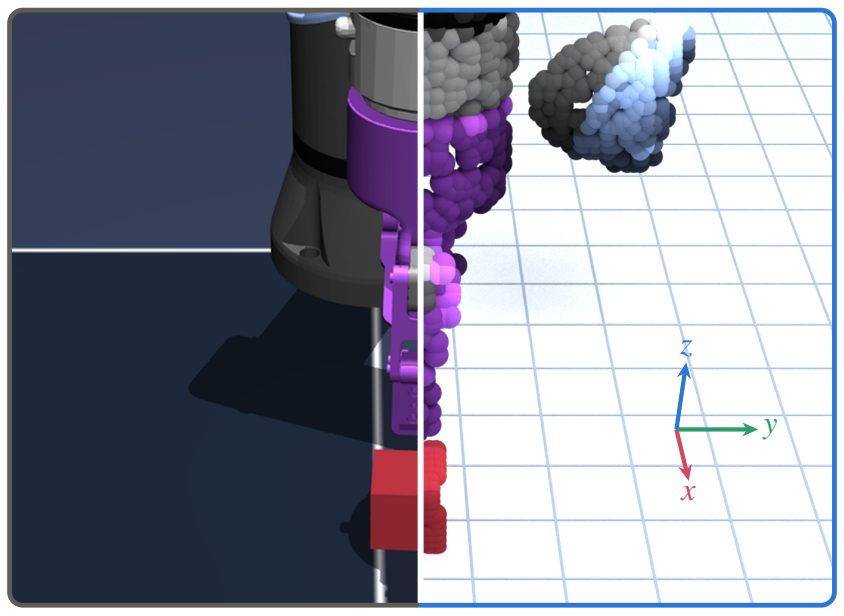

\looseness=1 JEPA4R observes the scene through one fixed depth camera (\cref{fig:obs}), as in a deployed cell. It is the eye-to-hand camera that renders each benchmark's pixels, treated as a time-of-flight sensor. Each depth image is lifted to world-frame points and cropped to a workspace box. The renderer's segmentation removes the table, floor, static fixtures and invisible goal markers and labels the task objects, which on Adroit Relocate include the visible goal marker. Merging points within 5\,mm voxels keeps surfaces near the camera from dominating. We then sample $N=1{,}200$ points. Each point stores xyz and rgb, so $o_t=P_t\in\mathbb{R}^{N\times6}$, normalized by a fixed affine map shared with the real arm.

\subsection{Architecture}
\label{sec:method-arch}

A point set has no pixel grid for a ViT to patchify, so JEPA4R's default encoder is the PointNet~\cite{qi2017pointnet} variant of DexArt~\cite{bao2023dexart}: a shared per-point MLP ($6{\to}1{,}080{\to}4{,}320$, GELU), an order-invariant max-pool and a linear head to 192 dimensions. Widened to 5.51M parameters, it matches the 5.50M ViT-tiny of \cref{sec:prelim}. With the projectors, predictor and action encoder of \cref{sec:prelim}, JEPA4R has about 18.0M parameters. It trains with \cref{eq:objective} under the recipe of \cref{sec:protocol} (10 epochs, batch 128, AdamW, bf16) on stable-worldmodel~\cite{maes2026stableworldmodel}.

\newsavebox{\jleftOG}\newsavebox{\jcapOG}\newlength{\jh}
\begin{table*}[t]
\sbox{\jleftOG}{\begin{minipage}[t]{0.56\linewidth}\vspace{0pt}%
\caption{\textbf{Evaluation on OGBench (success, \%).} We train a policy on the frozen latents of each encoder. We evaluate every encoder under the same protocol.}
\label{tab:ogbench_il}
\centering\footnotesize
\setlength{\tabcolsep}{3pt}
\begin{tabular}{@{}l*{7}{c}@{}}
\toprule
 & \multicolumn{3}{c}{Cube} & Scene & \multicolumn{3}{c}{Puzzle} \\
\cmidrule(lr){2-4}\cmidrule(lr){5-5}\cmidrule(l){6-8}
Encoder & Single & Double & Triple &  & $3\times3$ & $4\times4$ & $4\times5$ \\
\midrule
PLDM~\cite{sobal2025pldm} & 46.8 & 3.6 & 4.8 & 46.1 & \textbf{100.0} & 95.6 & 21.6 \\
LeJEPA~\cite{balestriero2025lejepa} & 0.0 & 0.0 & 0.0 & 41.1 & 96.8 & 99.2 & \textbf{96.4} \\
VISReg~\cite{wu2026visreg} & 1.2 & 0.0 & 0.0 & 40.5 & 1.2 & 0.4 & 0.0 \\
V-JEPA 2.1~\cite{murlabadia2026vjepa21} & 6.0 & 0.4 & 0.0 & 38.4 & \textbf{100.0} & \textbf{99.6} & 32.8 \\
LeWM~\cite{maes2026lewm} & 24.4 & 1.2 & 0.0 & 40.8 & \textbf{100.0} & \textbf{99.6} & 26.4 \\
\midrule
\textbf{JEPA4R (Ours)} & \textbf{78.4} & \textbf{41.6} & \textbf{18.4} & \textbf{64.8} & \textbf{100.0} & 95.6 & 68.0 \\
\bottomrule
\end{tabular}%
\end{minipage}}%
\sbox{\jcapOG}{\parbox{\dimexpr\linewidth-\wd\jleftOG-1.5em\relax}{\small Figure 4. Success of policies on frozen latents (paired data): OGBench Cube and Scene, MetaWorld and Adroit Relocate.}}%
\setlength{\jh}{\dimexpr\ht\jleftOG+\dp\jleftOG-\ht\jcapOG-\dp\jcapOG-8pt\relax}%
\noindent\usebox{\jleftOG}\hfill
\begin{minipage}[t]{\dimexpr\linewidth-\wd\jleftOG-1.5em\relax}\vspace{0pt}
\captionsetup{position=top}
\captionof{figure}{Success of policies on frozen latents (paired data): OGBench Cube and Scene, MetaWorld and Adroit Relocate.}
\label{fig:ogbench_plots}
\centering
\includegraphics[width=\linewidth,height=\jh,keepaspectratio]{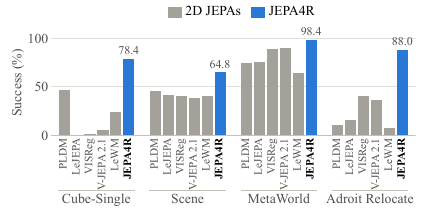}
\end{minipage}
\end{table*}
\newsavebox{\jtabCB}\newsavebox{\jleftCB}\newsavebox{\jcapCB}\newlength{\jhCB}
\begin{table*}[t]
\sbox{\jtabCB}{{\small\setlength{\tabcolsep}{3pt}%
\begin{tabular}[t]{@{}lccccc>{\columncolor{gray!8}[\tabcolsep][\tabcolsep]}cc@{}}
\toprule
 & \multicolumn{6}{c}{MetaWorld} & Adroit \\
\cmidrule(lr){2-7}\cmidrule(l){8-8}
Encoder & Assembly & Bin-pick & Button & Drawer & Hammer & Mean & Relocate \\
\midrule
Random (ViT-B), frozen & 0.0 & 0.0 & 2.7 & 0.0 & 0.0 & 0.5 & 0.0 \\
R3M (RN-50)~\cite{nair2022r3m} & \textbf{97.3} & 93.3 & 89.3 & \textbf{100.0} & \textbf{100.0} & 96.0 & 65.3 \\
MVP (ViT-B)~\cite{xiao2022mvp,radosavovic2022mvp} & 92.0 & 73.3 & 92.0 & \textbf{100.0} & 98.7 & 91.2 & 26.7 \\
VC-1 (ViT-B)~\cite{majumdar2023vc1} & 88.0 & 78.7 & 82.7 & \textbf{100.0} & 98.7 & 89.6 & 32.0 \\
VC-1 (ViT-L)~\cite{majumdar2023vc1} & 88.0 & 84.0 & 80.0 & \textbf{100.0} & 92.0 & 88.8 & 38.7 \\
CLIP (ViT-B)~\cite{radford2021clip} & 70.7 & 68.0 & 48.0 & \textbf{100.0} & 90.7 & 75.5 & 22.7 \\
STP (ViT-B)~\cite{yang2026stp} & 90.7 & 80.0 & 89.3 & \textbf{100.0} & \textbf{100.0} & 92.0 & 26.7 \\
\midrule
PLDM (ViT-Tiny)~\cite{sobal2025pldm} & 54.7 & 53.3 & 64.0 & \textbf{100.0} & \textbf{100.0} & 74.4 & 10.7 \\
LeJEPA (ViT-Tiny)~\cite{balestriero2025lejepa} & 34.7 & 61.3 & 78.7 & \textbf{100.0} & \textbf{100.0} & 74.9 & 16.0 \\
VISReg (ViT-B)~\cite{wu2026visreg} & 89.3 & 74.7 & 77.3 & \textbf{100.0} & \textbf{100.0} & 88.3 & 40.0 \\
V-JEPA 2 (ViT-L)~\cite{assran2025vjepa2} & 96.0 & 80.0 & 84.0 & \textbf{100.0} & 92.0 & 90.4 & -- \\
V-JEPA 2.1 (ViT-B)~\cite{murlabadia2026vjepa21} & 90.7 & 82.7 & 77.3 & \textbf{100.0} & \textbf{100.0} & 90.1 & 36.0 \\
LeWM (ViT-Tiny)~\cite{maes2026lewm} & 72.0 & 77.3 & 29.3 & 52.0 & 89.3 & 64.0 & 8.0 \\
\midrule
\textbf{JEPA4R (Ours)} & 96.0 & \textbf{100.0} & \textbf{98.7} & \textbf{100.0} & 97.3 & \textbf{98.4} & \textbf{88.0} \\
\bottomrule
\end{tabular}}}%
\sbox{\jleftCB}{\begin{minipage}[t]{\wd\jtabCB}\vspace{0pt}%
\caption{\textbf{Evaluation on CortexBench (success, \%).} MetaWorld (five tasks) and Adroit Relocate from CortexBench~\cite{majumdar2023vc1}. A policy is trained on the frozen latents of each encoder. Random (\mbox{ViT-B}) is a randomly initialized \mbox{ViT-B} that is never trained. Published numbers: the VC-1 study~\cite{majumdar2023vc1}, STP as reported by SPA~\cite{zhu2025spa} and V-JEPA~2 as reported by JEPA-VLA~\cite{miao2026jepavla} (no Adroit result, --). All other rows are our runs under the same protocol. Bold: best per column, ties included.}
\label{tab:metaworld}\label{tab:adroit}
\usebox{\jtabCB}%
\end{minipage}}%
\sbox{\jcapCB}{\parbox{\dimexpr\linewidth-\wd\jtabCB-1.5em\relax}{\small Figure 5. MetaWorld mean \vs\ Adroit Relocate success of each encoder in Tab.~2. Random and V-JEPA~2 (no Adroit result) are left out.}}%
\setlength{\jhCB}{\dimexpr\ht\jleftCB+\dp\jleftCB-\ht\jcapCB-\dp\jcapCB-8pt\relax}%
\noindent\usebox{\jleftCB}\hfill
\begin{minipage}[t]{\dimexpr\linewidth-\wd\jtabCB-1.5em\relax}\vspace{0pt}
\captionsetup{position=top}
\captionof{figure}{MetaWorld mean \vs\ Adroit Relocate success of each encoder in \cref{tab:metaworld}. Random and V-JEPA~2 (no Adroit result) are left out.}
\label{fig:cortex}
\centering
\includegraphics[width=\linewidth,height=\jhCB,keepaspectratio]{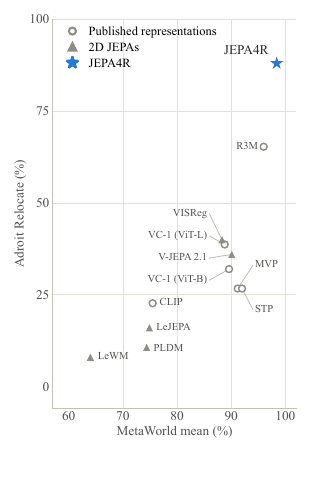}
\end{minipage}
\end{table*}

\subsection{Acting through a policy head}
\label{sec:method-policy}

\looseness=1 A world model helps a robot through the actions its latents support, so JEPA4R acts through a policy head $\pi_\omega$ on its frozen latents (\cref{fig:method}, top). The head reads the latents of recent frames, plus proprioception or a goal latent $z_g=f_\theta(o_g)$ where the task provides them, and outputs the next action or action block. We train $\pi_\omega$ while $f_\theta$ stays frozen and report the success rate of the trained policy (\cref{sec:supp-impl}). The head's success thus measures what the latents make available to control, and JEPA4R and every baseline we evaluate share the same heads. Fed with imagined latents, a latent-only head also lets JEPA4R look ahead: $g_\phi$ rolls the head's action blocks forward, and decoding the imagined rollout with $D_\psi$ (\cref{sec:method-decoder}) previews the policy in 4D point clouds before it acts.

\subsection{Latent-space reconstruction}
\label{sec:method-decoder}

To show what the world model imagines, a decoder $D_\psi$ maps every frozen latent $z_t$, observed or imagined, to its frame's 1,200-point cloud in one pass (\cref{fig:method}, bottom; \cref{fig:recon}). We train $D_\psi$ post hoc on the frozen world model, because a training-time reconstruction loss does not help the control of JEPA world models~\cite{maes2026lewm}.

\looseness=1 Expanding one 192-d vector into 1,200 metric points needs a strong geometric prior, so the decoder builds on PAGE-4D~\cite{zhou2026page4d}, a feed-forward reconstruction model for dynamic scenes. PAGE-4D combines a DINOv2 ViT-L~\cite{oquab2024dinov2}, 24 alternating frame and global attention layers and a DPT point-map head~\cite{ranftl2021dpt}. A set generator turns $z_t$ into PAGE-4D's input tokens: 256 learnable queries pass through three residual MLP blocks that $z_t$ modulates through FiLM~\cite{perez2018film}, and the resulting $16\times16$ grid of 1{,}024-d tokens enters PAGE-4D after its patch embedding. A learnable affine map moves the $224\times224$ point map into the target frame, an rgb branch colors it, and 1,200 pixels form $\hat P_t$. The decoder minimizes
\begin{equation}
  \mathcal{L}_{\mathrm{dec}}=\mathrm{CD}(\hat P_t,P_t)+\mathrm{SW}_2^2(\hat P_t,P_t)+0.1\,\mathcal{L}_{\mathrm{rgb}}(\hat P_t,P_t),
  \label{eq:decoder}
\end{equation}
\looseness=0 where $\hat P_t=D_\psi(z_t)$, CD is the symmetric Chamfer distance and $\mathrm{SW}_2^2$ the squared sliced Wasserstein-2 distance~\cite{bonneel2015sliced}, both on xyz, and $\mathcal{L}_{\mathrm{rgb}}$ is the L1 color error of nearest-neighbor matched points.

The decoder trains on the OGBench Cube data, with latents from the frozen epoch-10 JEPA4R model, in two stages that follow the linear-probe-then-fine-tune (LP-FT) order~\cite{kumar2022lpft}. Stage A trains the 5.75M new parameters with PAGE-4D (about 0.94B) frozen. Stage B adds rank-4 LoRA~\cite{hu2022lora} adapters (0.88M) to the query and value projections of all 72 attention blocks, which lowers the validation error from 0.87 to 0.79\,cm. For imagined rollouts, $D_\psi$ decodes every latent that $g_\phi$ predicts from the context frames at environment steps 0, 5 and 10, which yields a decoded 4D rollout.

\section{Experiments}
\label{sec:exp}

\begin{figure*}[t]
\centering
\definecolor{jmann}{HTML}{E8730C}%
\scalebox{0.997}{\begin{tikzpicture}[x=1cm, y=1cm, font=\footnotesize,
    ann/.style={draw=jmann, line width=0.6pt},
    box/.style={rounded corners=1pt},
    steplbl/.style={anchor=north, inner sep=0, font=\fontsize{6}{7}\selectfont, text=black!75}]
  \node[rotate=90] at (0.275,-1.080) {Key frames};
  \node[anchor=base west, inner sep=0] at (0.550,-0.310) {Cube-Double};
  \node[anchor=south west, inner sep=0] at (0.550,-1.780) {\includegraphics[width=1.312cm,height=1.400cm]{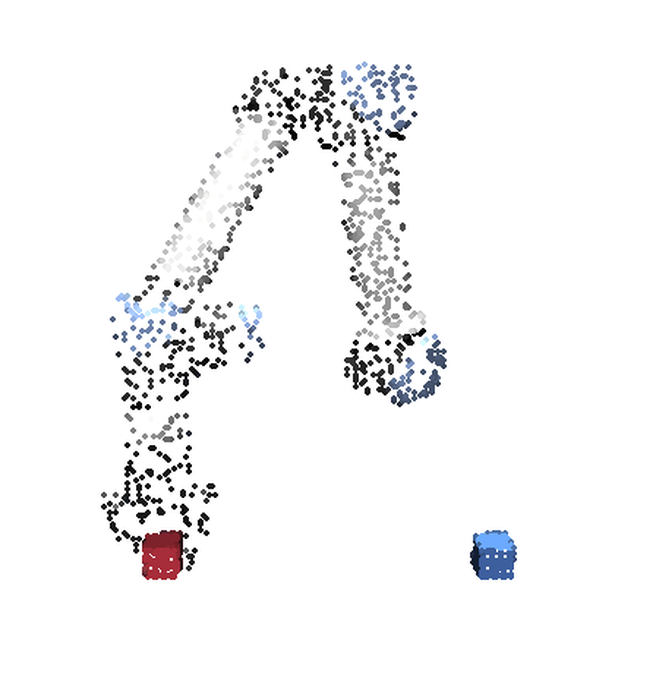}};
  \node[anchor=south west, inner sep=0] at (1.943,-1.780) {\includegraphics[width=1.312cm,height=1.400cm]{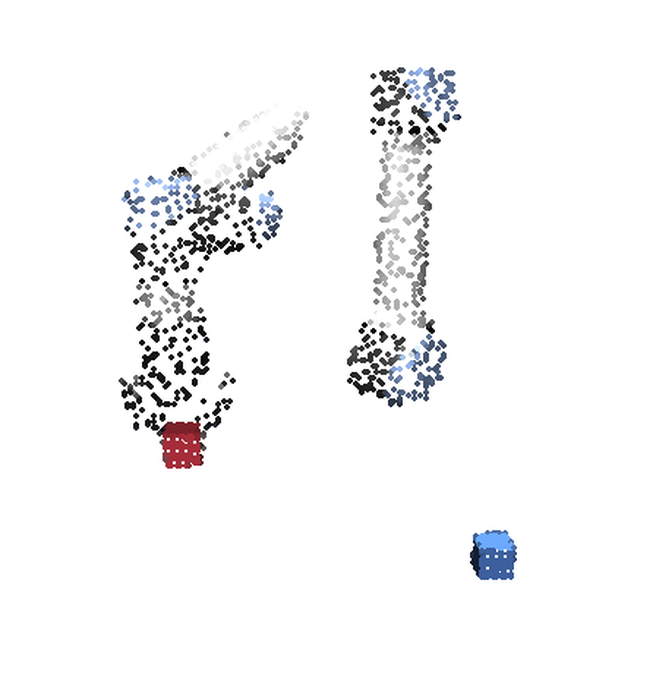}};
  \node[anchor=south west, inner sep=0] at (3.335,-1.780) {\includegraphics[width=1.312cm,height=1.400cm]{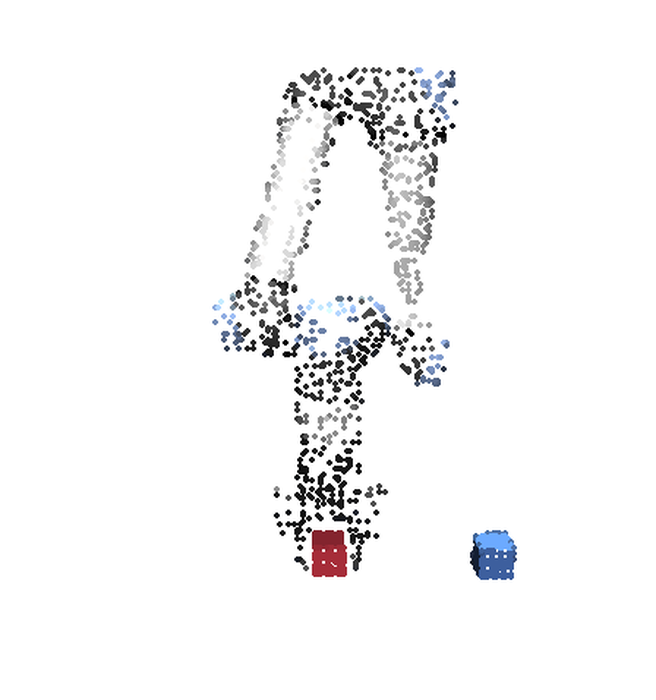}};
  \node[anchor=south west, inner sep=0] at (4.728,-1.780) {\includegraphics[width=1.312cm,height=1.400cm]{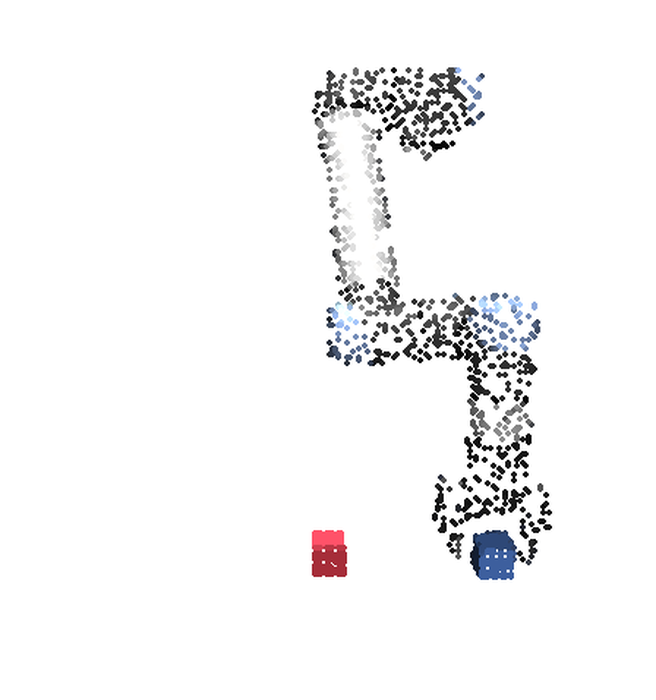}};
  \node[anchor=south west, inner sep=0] at (6.120,-1.780) {\includegraphics[width=1.312cm,height=1.400cm]{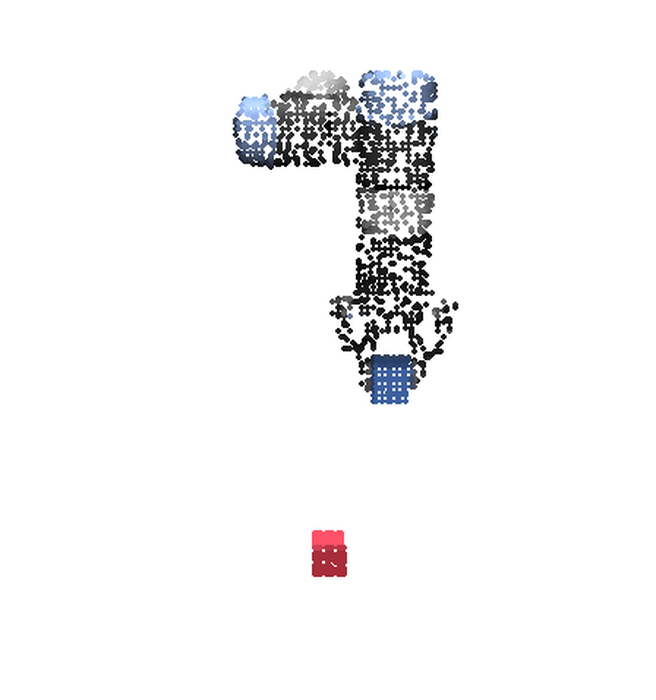}};
  \node[anchor=south west, inner sep=0] at (7.513,-1.780) {\includegraphics[width=1.312cm,height=1.400cm]{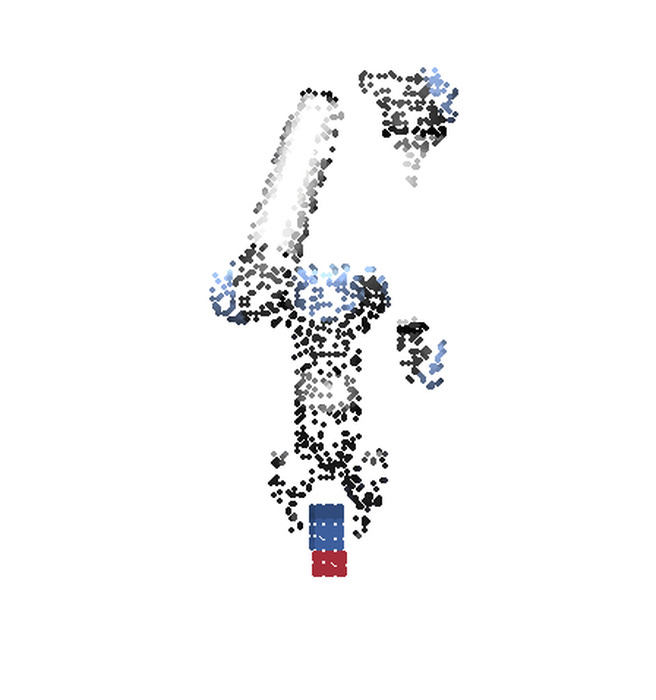}};
  \node[anchor=base west, inner sep=0] at (9.075,-0.310) {Scene};
  \node[anchor=south west, inner sep=0] at (9.075,-1.780) {\includegraphics[width=1.312cm,height=1.400cm]{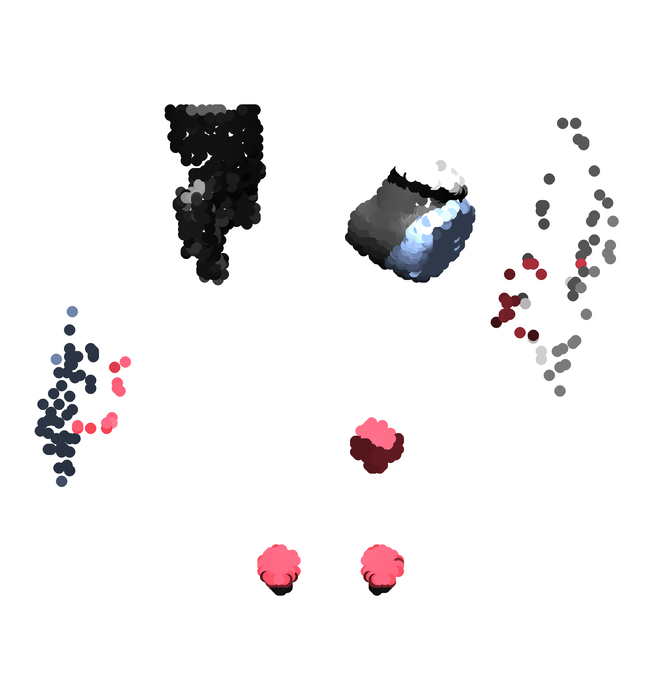}};
  \node[anchor=south west, inner sep=0] at (10.468,-1.780) {\includegraphics[width=1.312cm,height=1.400cm]{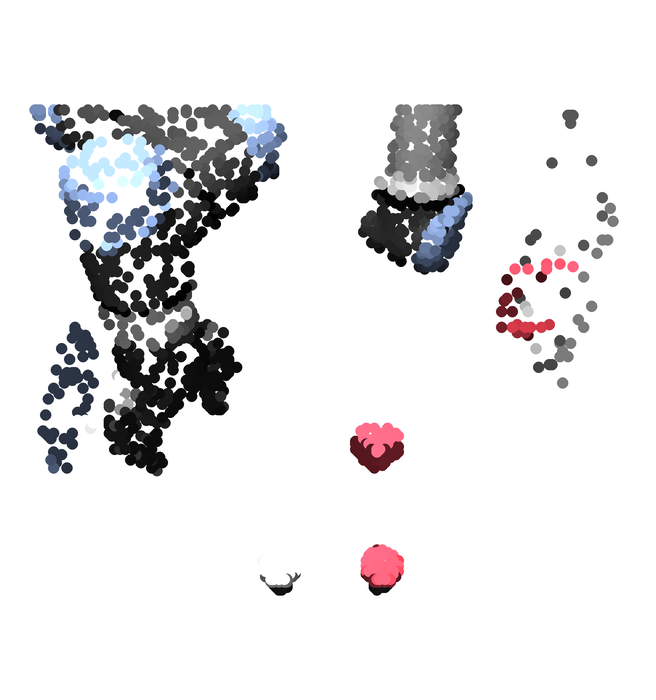}};
  \node[anchor=south west, inner sep=0] at (11.860,-1.780) {\includegraphics[width=1.312cm,height=1.400cm]{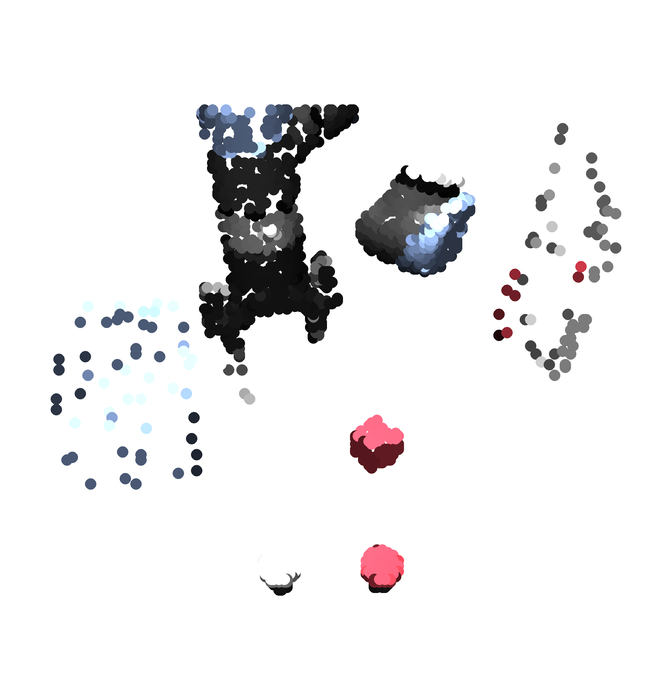}};
  \node[anchor=south west, inner sep=0] at (13.253,-1.780) {\includegraphics[width=1.312cm,height=1.400cm]{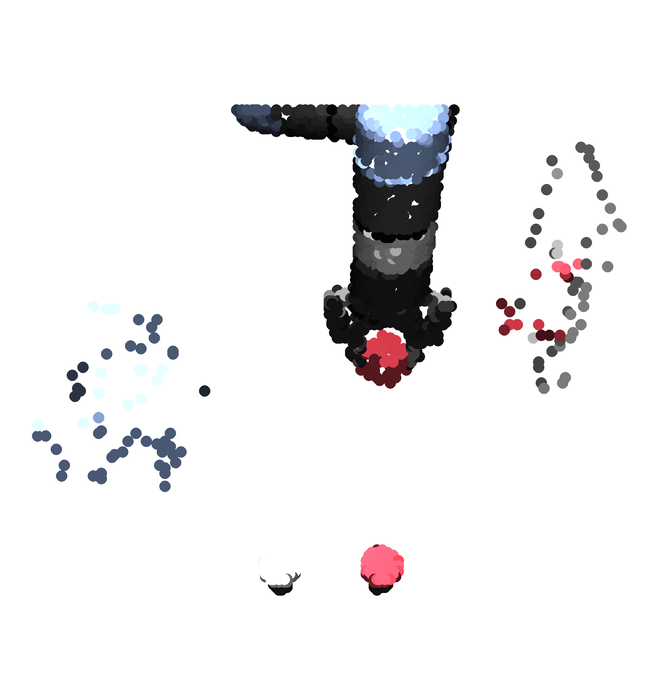}};
  \node[anchor=south west, inner sep=0] at (14.645,-1.780) {\includegraphics[width=1.312cm,height=1.400cm]{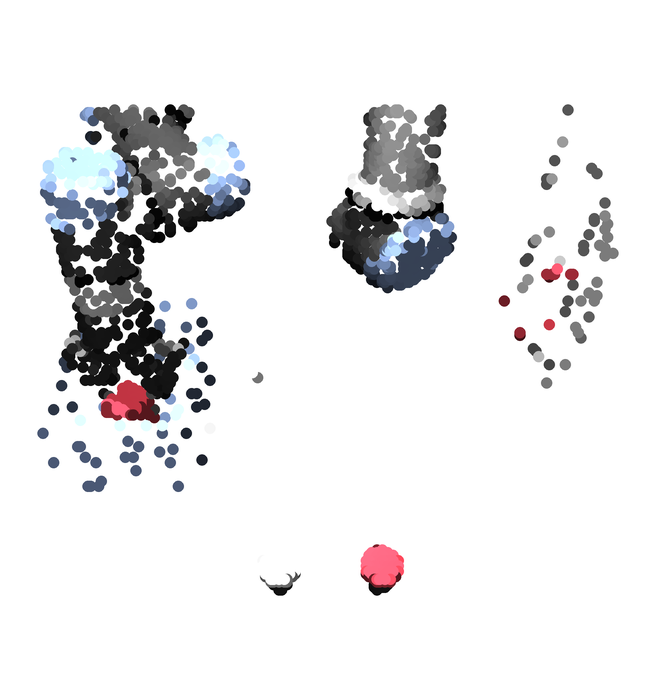}};
  \node[anchor=south west, inner sep=0] at (16.038,-1.780) {\includegraphics[width=1.312cm,height=1.400cm]{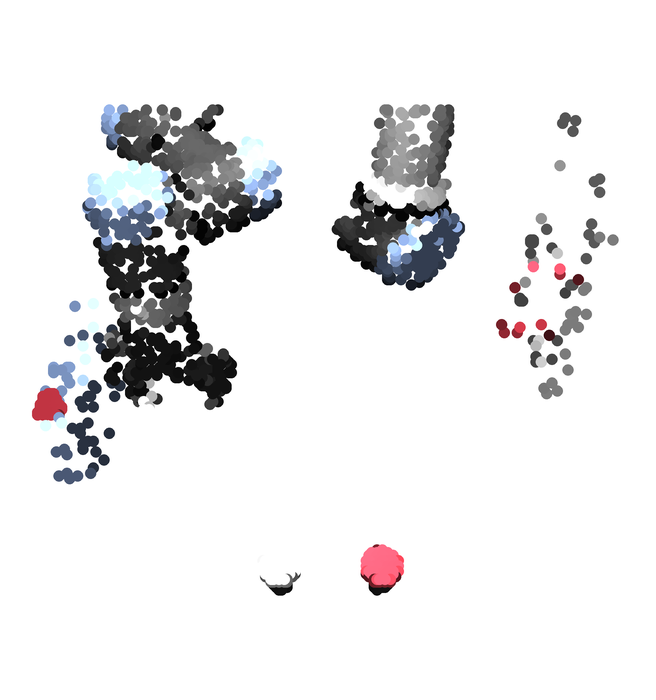}};
  \node[rotate=90] at (0.275,-3.100) {Key frames};
  \node[anchor=base west, inner sep=0] at (0.550,-2.330) {Puzzle};
  \node[anchor=south west, inner sep=0] at (0.550,-3.800) {\includegraphics[width=1.312cm,height=1.400cm]{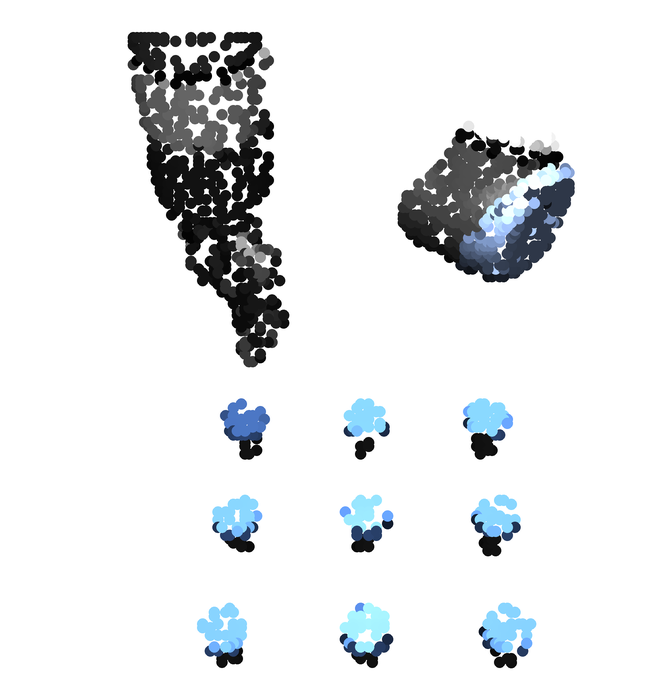}};
  \node[anchor=south west, inner sep=0] at (1.943,-3.800) {\includegraphics[width=1.312cm,height=1.400cm]{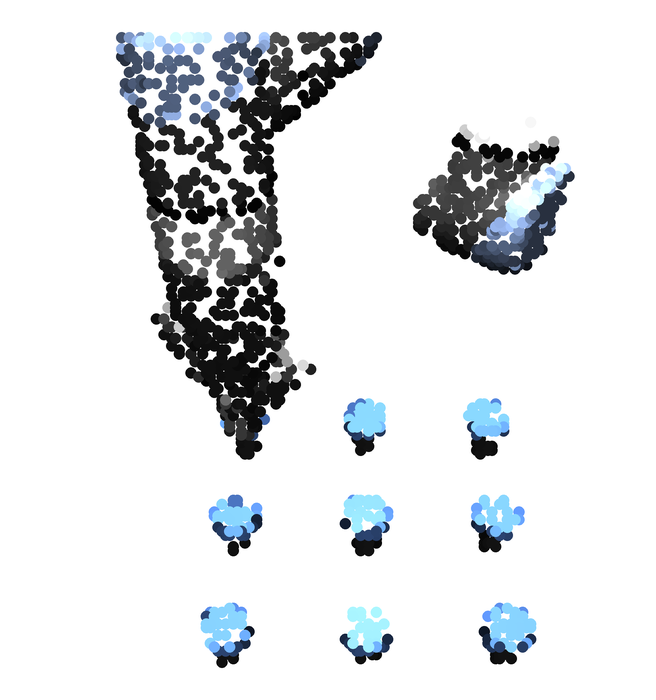}};
  \node[anchor=south west, inner sep=0] at (3.335,-3.800) {\includegraphics[width=1.312cm,height=1.400cm]{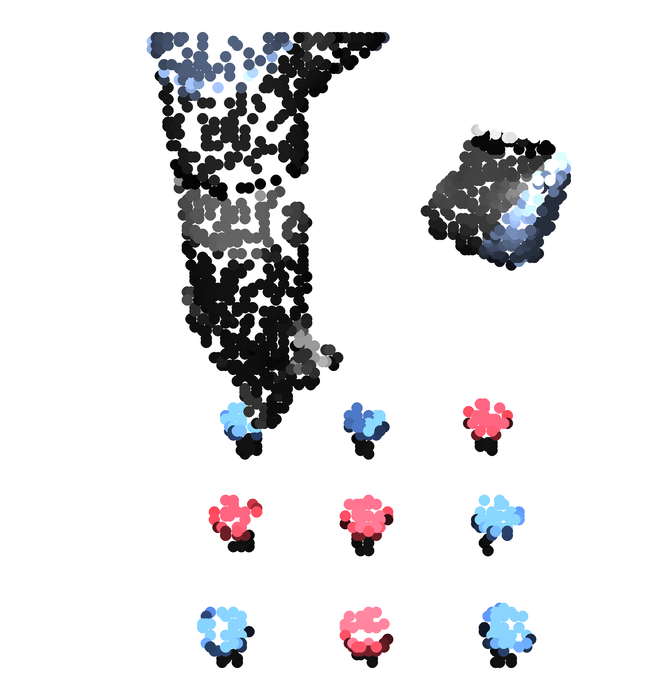}};
  \node[anchor=south west, inner sep=0] at (4.728,-3.800) {\includegraphics[width=1.312cm,height=1.400cm]{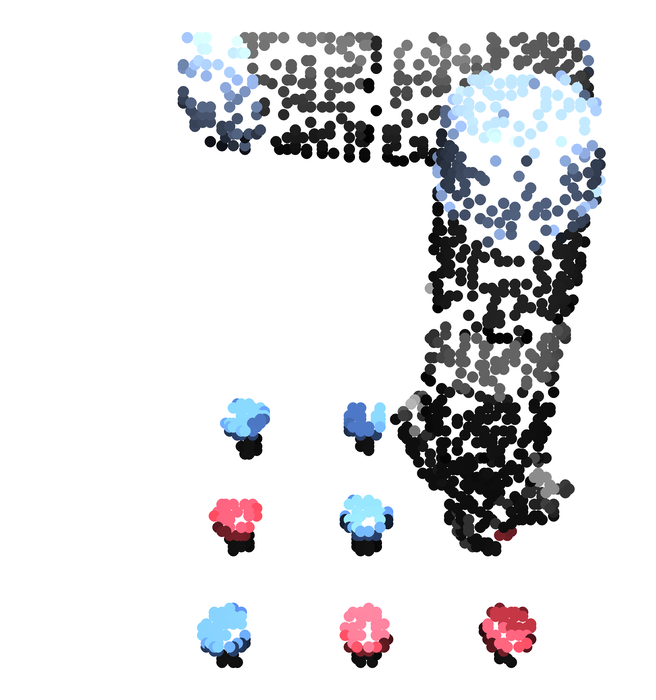}};
  \node[anchor=south west, inner sep=0] at (6.120,-3.800) {\includegraphics[width=1.312cm,height=1.400cm]{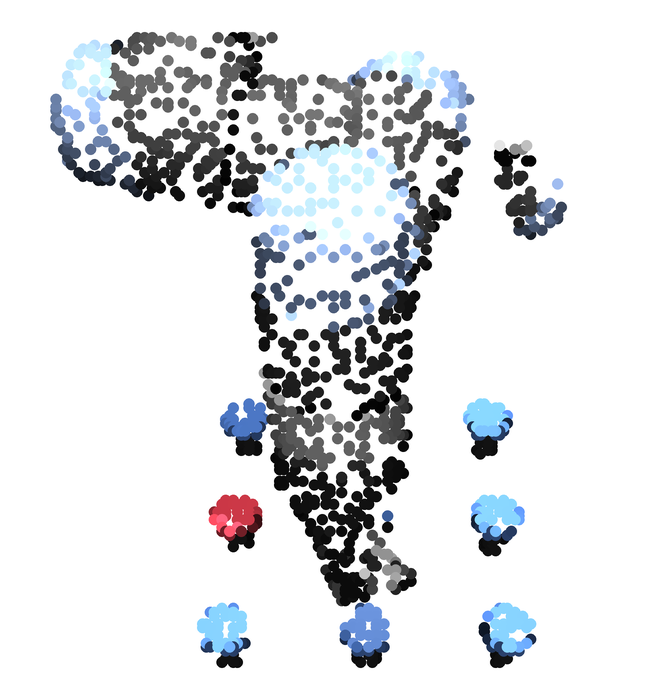}};
  \node[anchor=south west, inner sep=0] at (7.513,-3.800) {\includegraphics[width=1.312cm,height=1.400cm]{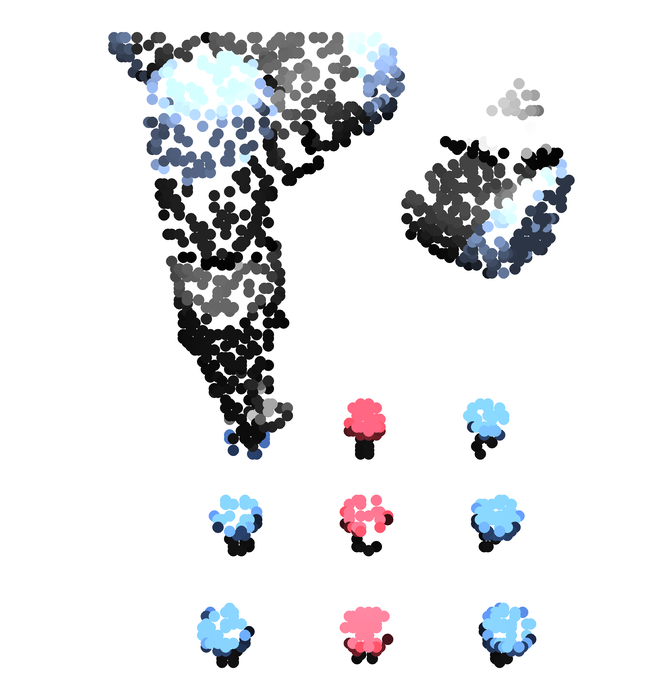}};
  \node[anchor=base west, inner sep=0] at (9.075,-2.330) {MetaWorld Assembly};
  \node[anchor=south west, inner sep=0] at (9.075,-3.800) {\includegraphics[width=1.312cm,height=1.400cm]{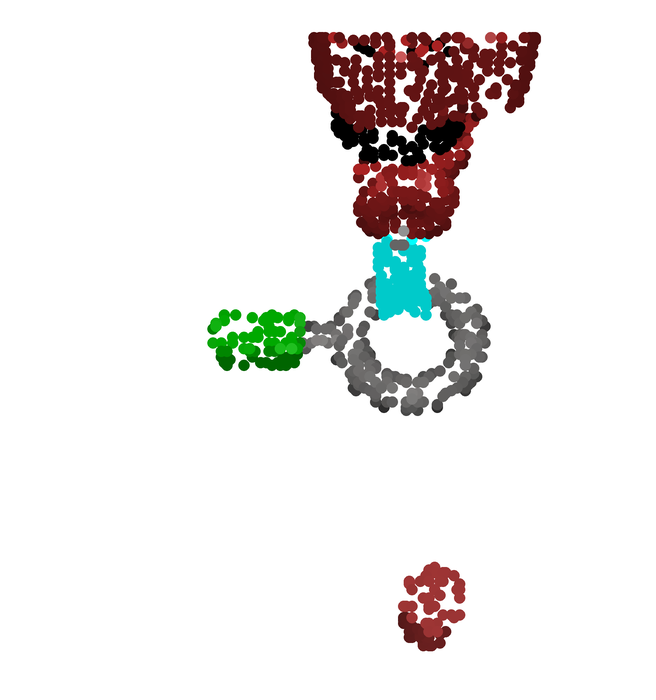}};
  \node[anchor=south west, inner sep=0] at (10.468,-3.800) {\includegraphics[width=1.312cm,height=1.400cm]{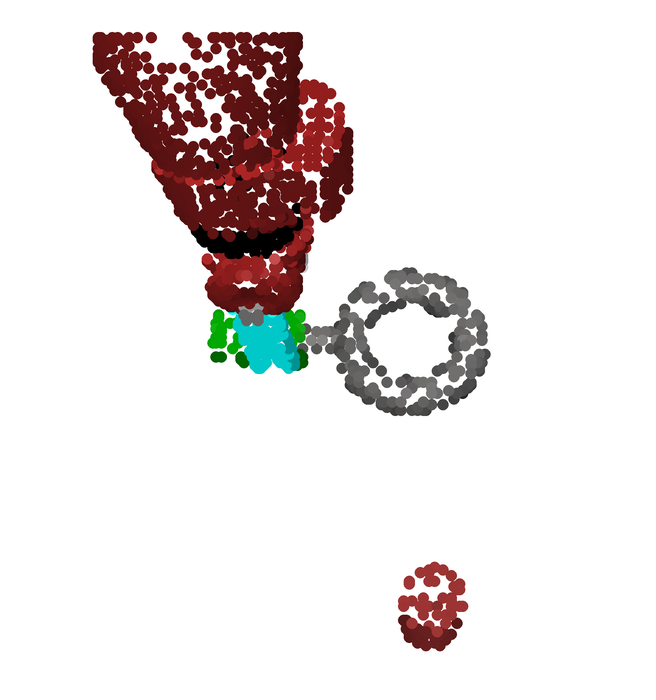}};
  \node[anchor=south west, inner sep=0] at (11.860,-3.800) {\includegraphics[width=1.312cm,height=1.400cm]{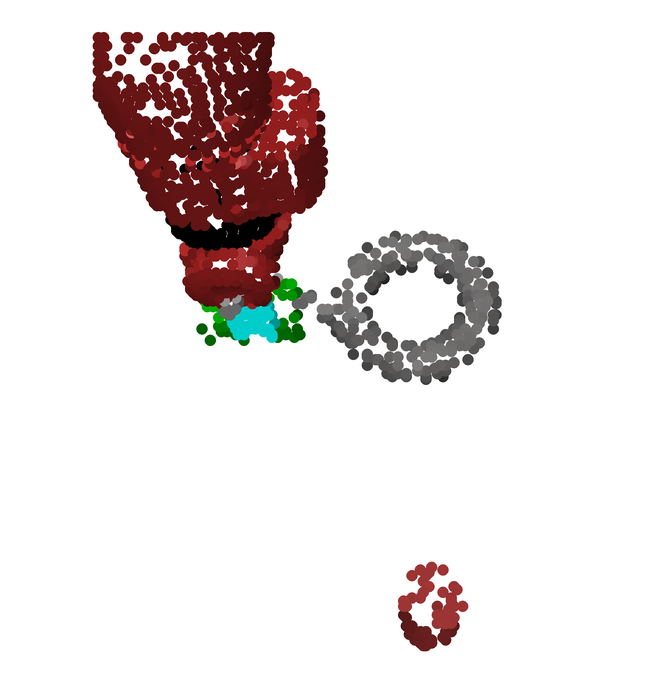}};
  \node[anchor=south west, inner sep=0] at (13.253,-3.800) {\includegraphics[width=1.312cm,height=1.400cm]{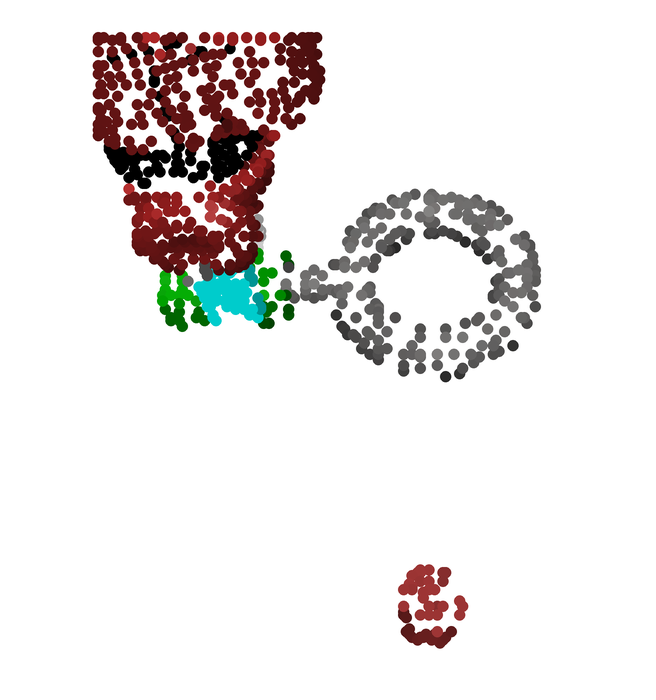}};
  \node[anchor=south west, inner sep=0] at (14.645,-3.800) {\includegraphics[width=1.312cm,height=1.400cm]{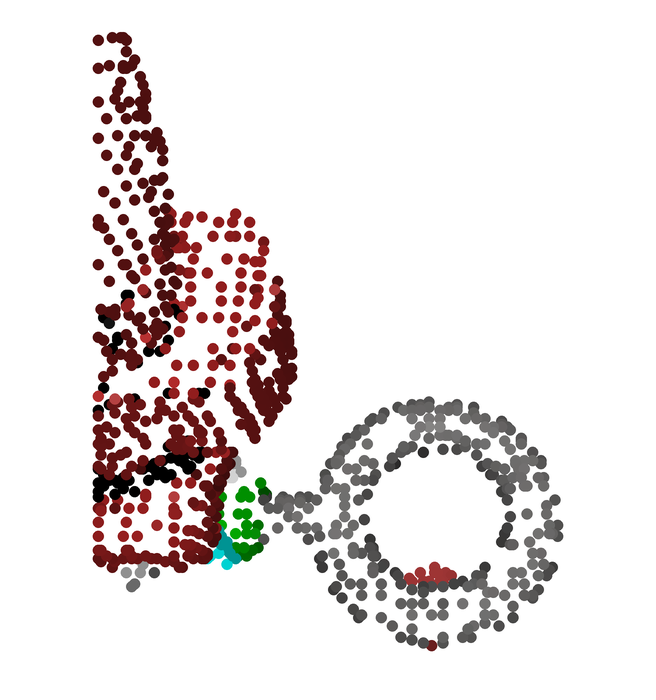}};
  \node[anchor=south west, inner sep=0] at (16.038,-3.800) {\includegraphics[width=1.312cm,height=1.400cm]{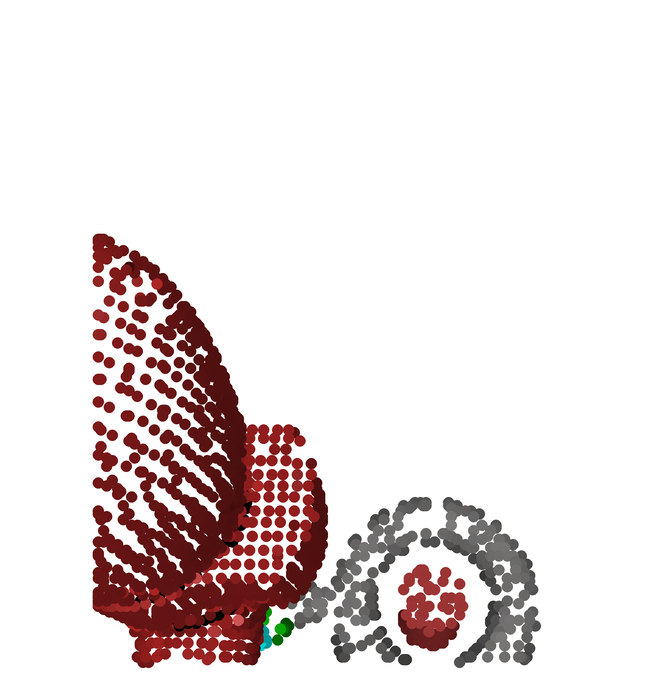}};
  \node[steplbl] at (1.206,-1.790) {start};
  \node[steplbl] at (2.599,-1.790) {red lifted};
  \node[steplbl] at (3.991,-1.790) {red placed};
  \node[steplbl] at (5.384,-1.790) {blue grasped};
  \node[steplbl] at (6.776,-1.790) {blue lifted};
  \node[steplbl] at (8.169,-1.790) {stacked};
  \node[steplbl] at (9.731,-1.790) {start};
  \node[steplbl] at (11.124,-1.790) {button pressed};
  \node[steplbl] at (12.516,-1.790) {drawer open};
  \node[steplbl] at (13.909,-1.790) {cube grasped};
  \node[steplbl] at (15.301,-1.790) {cube moved};
  \node[steplbl] at (16.694,-1.790) {cube in drawer};
  \node[steplbl] at (1.206,-3.810) {all off};
  \node[steplbl] at (2.599,-3.810) {all off};
  \node[steplbl] at (3.991,-3.810) {4 lit};
  \node[steplbl] at (5.384,-3.810) {3 lit};
  \node[steplbl] at (6.776,-3.810) {1 lit};
  \node[steplbl] at (8.169,-3.810) {3 lit};
  \node[steplbl] at (9.731,-3.810) {start};
  \node[steplbl] at (11.124,-3.810) {handle grasped};
  \node[steplbl] at (12.516,-3.810) {ring lifted};
  \node[steplbl] at (13.909,-3.810) {ring moved};
  \node[steplbl] at (15.301,-3.810) {ring lowered};
  \node[steplbl] at (16.694,-3.810) {ring on peg};
  \draw[ann, box] (2.251,-1.330) rectangle (2.359,-1.208);
  \draw[ann, box] (3.943,-1.548) rectangle (4.043,-1.424);
  \draw[ann, box] (5.658,-1.554) rectangle (5.776,-1.424);
  \draw[ann, box] (6.846,-1.202) rectangle (6.962,-1.080);
  \draw[ann, box] (8.115,-1.554) rectangle (8.221,-1.372);
  \draw[ann, box] (10.972,-1.560) rectangle (11.078,-1.456);
  \draw[ann, box] (11.932,-1.364) rectangle (12.296,-0.956);
  \draw[ann, box] (13.963,-1.130) rectangle (14.083,-1.024);
  \draw[ann, box] (14.831,-1.230) rectangle (14.953,-1.124);
  \draw[ann, box] (16.086,-1.236) rectangle (16.186,-1.140);
  \draw[ann, box] (10.866,-3.168) rectangle (11.094,-2.988);
  \draw[ann, box] (12.428,-3.184) rectangle (12.878,-2.846);
  \draw[ann, box] (13.797,-3.178) rectangle (14.349,-2.762);
  \draw[ann, box] (15.145,-3.712) rectangle (15.785,-3.176);
  \draw[ann, box] (16.598,-3.750) rectangle (17.134,-3.378);
\end{tikzpicture}}
\caption{Key frames in simulation: OGBench Cube-Double, Scene and Puzzle, and MetaWorld Assembly. The label under each frame gives the state it shows, and the orange box marks the object that changed.}
\label{fig:sim}
\end{figure*}
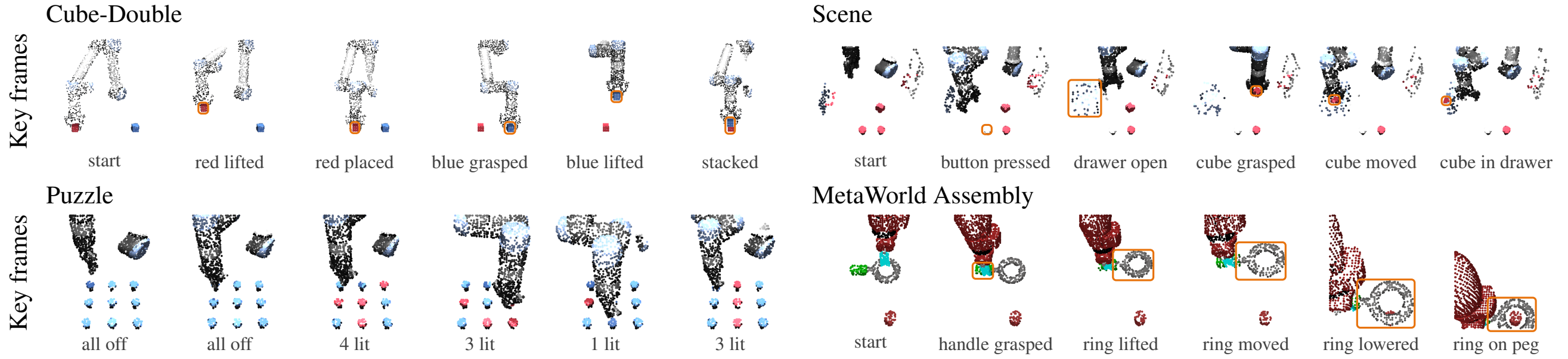

\subsection{Benchmarks and Metrics}
\paragraph{OGBench.}
We use the Cube (Single, Double, Triple, Quadruple), Scene and Puzzle ($3\times3$, $4\times4$, $4\times5$) environments of OGBench~\cite{park2025ogbench} (\cref{tab:ogbench_il}, \cref{fig:sim}). Cube and Scene are 3D manipulation tasks, and Puzzle comes in three board sizes. The metric is the \textit{success rate} on each environment's five official evaluation goals.

\paragraph{CortexBench.}
We use five MetaWorld~\cite{yu2019metaworld} tasks (Assembly, Bin-pick, Button, Drawer and Hammer) and the Adroit~\cite{rajeswaran2018dapg} dexterous-hand task Relocate (\cref{tab:metaworld}). The metric is the \textit{success rate}, taken as the best of 20 evaluations. On MetaWorld, the published representations of the VC-1 study~\cite{majumdar2023vc1} use the same metric.

\begin{table}[t]
\centering
\caption{\textbf{Decoding frozen latents into point clouds} (OGBench Cube, cm). (a)~Per frame: mean, median and 90th percentile (p90) of the bidirectional nearest-neighbor (NN) distance between decoded and observed points; Outl.: share of points with an NN distance above 3\,cm; Cube: cube position error. Decoding the latent of another frame instead gives 12.31\,cm. (b)~Open-loop imagination; decoded true latents decode encoder, not predicted, latents, and another episode's actions replay the actions of a different episode.}
\label{tab:recon}
\small
\setlength{\tabcolsep}{3.2pt}
\begin{tabular}{@{}lccccc@{}}
\multicolumn{6}{@{}l}{\textit{(a) Per-frame reconstruction}} \\
\toprule
Latent & Mean & Median & p90 & Outl.\,(\%) & Cube \\
\midrule
JEPA4R & 0.79 & 0.78 & 0.94 & 2.89 & 1.05 \\
\bottomrule
\end{tabular}

\vspace{6pt}
\begin{tabular}{@{}lccc@{}}
\multicolumn{4}{@{}l}{\textit{(b) Open-loop imagination}} \\
\toprule
NN distance at env step & 15 & 100 & 200 \\
\midrule
Decoded imagination & 0.75 & 0.85 & 0.89 \\
Decoded true latents & 0.75 & 0.80 & 0.74 \\
Another episode's actions & 1.86 & 13.23 & 11.46 \\
\midrule
Cube position, imagination & 0.92 & 1.10 & 1.39 \\
\bottomrule
\end{tabular}
\end{table}

\subsection{Implementation Details}
\label{sec:protocol}
Except on Puzzle, which follows the noisy recipe of OGBench, the world-model data mix three kinds of trajectories: noisy expert control, the same control interrupted by random actions, and purely random actions, with the mixture set per benchmark (\cref{sec:supp-impl}). The image and point-cloud datasets record the same simulator rollouts row by row, so states, actions, splits and starts coincide. Per task, we train JEPA4R on the point clouds (\cref{sec:method-arch}) and LeWM, PLDM and LeJEPA on the paired images, while VISReg and V-JEPA~2.1 use the authors' released weights (\cref{sec:supp-baselines}). We then train a policy on the frozen latents and report its success rate. All methods use the same policies and evaluation episodes. The JEPA4R world model trains with AdamW for 10 epochs, with a learning rate of $5{\times}10^{-5}$, a batch size of 128 and bf16 precision. All training runs on one NVIDIA RTX 5090 (32\,GB) with a 16-vCPU Intel Xeon Gold 6459C and 90\,GB of RAM, using PyTorch 2.8 and CUDA 12.8. For more details, see \cref{sec:supp-impl}.

\subsection{Point Clouds Outperform Pixels}
\label{sec:exp-imitation}

\paragraph{OGBench.}
\looseness=1 \textit{Cube} and \textit{Scene} are 3D manipulation tasks, and JEPA4R outperforms every other method on Cube-Single, Cube-Double, Cube-Triple and Scene (\cref{tab:ogbench_il}), with 78.4\% on Cube-Single and 64.8\% on Scene. On Cube-Quadruple, every encoder, JEPA4R included, reaches 0.0. Its goals ask the arm to move, stack or cycle four cubes, up to four pick-and-place steps each, and the policy must learn them from random play data~\cite{park2025ogbench}. This long-horizon policy is hard to learn from any representation: all six offline goal-conditioned RL methods of OGBench, GCIQL included, also reach 0 on it even with ground-truth states~\cite{park2025ogbench}. We therefore leave Cube-Quadruple out of \cref{tab:ogbench_il}. On \textit{Puzzle}, most methods come close to full success on $3\times3$ and $4\times4$, JEPA4R with 100.0\% and 95.6\%. On $4\times5$, JEPA4R reaches 68.0\%, above every other method except LeJEPA.

\newsavebox{\jabA}\newsavebox{\jabB}\newsavebox{\jabC}\newsavebox{\jabD}\newsavebox{\jgrid}\newlength{\jcolA}\newlength{\jcolB}
\begin{table*}[t]
\caption{\textbf{Ablations.} (a)~Decoder backbone (OGBench Cube, cm). NN: mean bidirectional nearest-neighbor distance between decoded and observed points, per frame; Cube: cube position error, per frame; Step 200: NN distance of the decoded imagination at environment step 200. (b)~Point-cloud encoder, (c)~temporal window and (d)~training hyperparameters (OGBench Scene, mean success over the five tasks, \%). Context frames are spaced five environment steps apart. Default: learning rate $5{\times}10^{-5}$, batch size 128. LR: learning rate; $^\dagger$:~default.}
\label{tab:ablations}
\sbox{\jabA}{\footnotesize\setlength{\tabcolsep}{3pt}%
\begin{tabular}[t]{@{}lccc@{}}
\toprule
Backbone & NN & Cube & Step 200 \\
\midrule
PAGE-4D~\cite{zhou2026page4d}$^\dagger$ & 0.79 & 1.05 & 0.89 \\
VGGT~\cite{wang2025vggt} & 0.79 & 1.01 & 0.88 \\
Depth Anything 3~\cite{lin2026da3} & 0.75 & 0.94 & 0.85 \\
\bottomrule
\end{tabular}}%
\sbox{\jabB}{\footnotesize\setlength{\tabcolsep}{3pt}%
\begin{tabular}[t]{@{}lc@{}}
\toprule
Encoder & Success \\
\midrule
PointNet~\cite{qi2017pointnet}$^\dagger$ & 64.8 \\
PointNet++~\cite{qi2017pointnetpp} & 60.5 \\
DGCNN~\cite{wang2019dgcnn} & 62.4 \\
PointPatchTransformer~\cite{gyenes2024pprl} & 57.1 \\
\bottomrule
\end{tabular}}%
\sbox{\jabC}{\footnotesize\setlength{\tabcolsep}{3pt}%
\begin{tabular}[t]{@{}cc@{}}
\toprule
Context frames & Success \\
\midrule
2 & 63.2 \\
3$^\dagger$ & 64.8 \\
4 & 64.8 \\
\bottomrule
\end{tabular}}%
\sbox{\jabD}{\footnotesize\setlength{\tabcolsep}{3pt}%
\begin{tabular}[t]{@{}lc@{}}
\toprule
Setting & Success \\
\midrule
Default & 64.8 \\
LR $2.5{\times}10^{-5}$ & 61.9 \\
LR $10^{-4}$ & 62.9 \\
Batch 64 & 57.6 \\
Batch 256 & 68.5 \\
\bottomrule
\end{tabular}}%
\setlength{\jcolA}{\wd\jabA}\ifdim\wd\jabB>\jcolA\setlength{\jcolA}{\wd\jabB}\fi
\setlength{\jcolB}{\wd\jabC}\ifdim\wd\jabD>\jcolB\setlength{\jcolB}{\wd\jabD}\fi
\sbox{\jgrid}{%
\begin{minipage}[t]{\jcolA}\vspace{0pt}
\begin{subtable}{\linewidth}
\centering
\caption{Decoder backbone}
\label{tab:abl-decoder}
\usebox{\jabA}
\end{subtable}\par\vspace{6pt}
\begin{subtable}{\linewidth}
\centering
\caption{Point-cloud encoder}
\label{tab:abl-encoder}
\usebox{\jabB}
\end{subtable}
\end{minipage}\hspace{12pt}%
\begin{minipage}[t]{\jcolB}\vspace{0pt}
\begin{subtable}{\linewidth}
\centering
\caption{Temporal window}
\label{tab:abl-time}
\usebox{\jabC}
\end{subtable}\par\vspace{6pt}
\begin{subtable}{\linewidth}
\centering
\caption{Hyperparameters}
\label{tab:abl-hparam}
\usebox{\jabD}
\end{subtable}
\end{minipage}}%
\noindent\usebox{\jgrid}\hfill
\begin{minipage}[t]{\dimexpr\linewidth-\wd\jgrid-1.5em\relax}\vspace{0pt}
\centering
\includegraphics[width=\linewidth,height=\dimexpr\ht\jgrid+\dp\jgrid\relax,keepaspectratio]{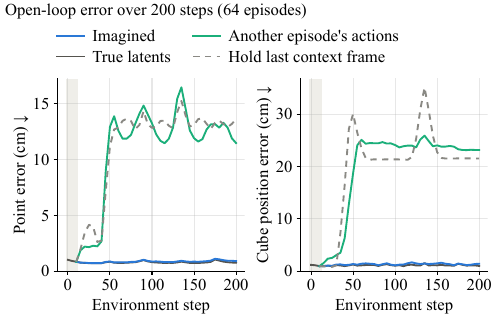}
\end{minipage}\par\vspace{4pt}
\captionof{figure}{\textbf{Imagined rollouts keep a mean point error within 1.1\,cm over 200 environment steps} (OGBench Cube, mean of 64 episodes). Point error (left) and cube position error (right); gray band: three context frames; \emph{True latents} decodes encoder, not predicted, latents. \Cref{fig:teaser}a shows one decoded 4D rollout.}
\label{fig:recon}
\end{table*}

\paragraph{CortexBench.}
\looseness=1 \textit{MetaWorld.}
JEPA4R reaches a mean success of 98.4 and ranks first (\cref{tab:metaworld}, \cref{fig:cortex}), above every published representation and every 2D JEPA. It is highest on Bin-pick, Button and Drawer, tying with most methods at 100.0 on Drawer. On Bin-pick it reaches 100.0, where the runner-up, R3M, reaches 93.3.
\textit{Adroit.}
JEPA4R reaches a success of 88.0 on Relocate and ranks first (\cref{tab:adroit}), above the runner-up, R3M, at 65.3, and above every 2D JEPA, the best of which reaches 40.0.

\vspace{\baselineskip}
\textit{2D pretraining does not close the gap.}
Among the 2D JEPAs, VISReg is pretrained on a large image collection, V-JEPA~2 and V-JEPA~2.1 on large video collections, and PLDM and LeJEPA are trained on the same paired data. All of them stay below JEPA4R wherever they are reported on Cube-Single, Cube-Double, Cube-Triple, Scene, the MetaWorld mean and Adroit Relocate (\cref{tab:ogbench_il,tab:metaworld}), so larger-scale pretraining in 2D does not substitute for a 3D observation.

\subsection{Latents Retain Geometry}
\label{sec:exp-recon}

\looseness=1 We decode frozen JEPA4R latents back to point clouds to see how much geometry they retain (OGBench Cube). Decoding single frames, the bidirectional mean nearest-neighbor distance between decoded and observed points is 0.79\,cm over 4,096 validation frames (\cref{tab:recon}). Letting the predictor roll forward autoregressively from the context frames and decoding every step, the mean point error of the imagined rollout stays within 1.1\,cm over 200 environment steps (\cref{fig:recon}). The world model never learns to reconstruct, so this decodability shows that the JEPA objective itself keeps the metric geometry of the scene in the latents, and it lets the imagined outcome be checked before the robot acts.

\subsection{Ablation Study}
\label{sec:exp-analysis}

\paragraph{Decoder backbone.}
\looseness=1 We replace PAGE-4D in the decoder with other feed-forward 3D reconstruction models, keep the set generator, the affine map, the training data and both training stages unchanged, and evaluate again on OGBench Cube (\cref{tab:abl-decoder}). PAGE-4D and VGGT give the same per-frame error (0.79\,cm), and Depth Anything~3 reaches 0.75\,cm. The gap stays as small on imagined rollouts: at environment step 200, PAGE-4D reaches 0.89\,cm, VGGT 0.88\,cm and Depth Anything~3 0.85\,cm. Although PAGE-4D extends VGGT to dynamic scenes, the two decode rollouts equally well. Across the three backbones, each error varies by at most 0.11\,cm, so the geometry in the frozen latents does not depend on one particular decoder.

\paragraph{Point-cloud encoder.}
We replace JEPA4R's PointNet with other point-cloud encoders of the same parameter count, keep the rest of the world model, the training data and the policy head unchanged, and compare on OGBench Scene (\cref{tab:abl-encoder}). PointNet reaches the highest success (64.8), so we keep it as JEPA4R's default encoder.

\paragraph{Temporal window.}
By default, the world model reads three context frames spaced five environment steps apart. We change the number of context frames to two and four, with the policy head unchanged, and compare on OGBench Scene (\cref{tab:abl-time}). With two frames, success is 63.2, and from three frames on it no longer changes (64.8 with three and with four), so the default of three frames is enough.

\paragraph{Training hyperparameters.}
JEPA4R trains with a learning rate of $5{\times}10^{-5}$ and a batch size of 128. We halve and double the learning rate to $2.5{\times}10^{-5}$ and $10^{-4}$, and set the batch size to 64 and 256 while keeping the learning rate and the number of epochs, on OGBench Scene (\cref{tab:abl-hparam}). JEPA4R stays between 57.6 and 68.5 in every setting. Changing the learning rate costs at most 2.9 points, whereas the batch size matters more, with 57.6 at 64 and 68.5 at 256.

\begin{figure*}[t]
\centering
\begin{tikzpicture}[x=1cm, y=1cm, font=\footnotesize]
  \node[anchor=south west, inner sep=0] at (0.550,-0.320) {Real RM75-B arm};
  \node[anchor=north west, inner sep=0] at (0.550,-0.380) {\includegraphics[width=16.8cm]{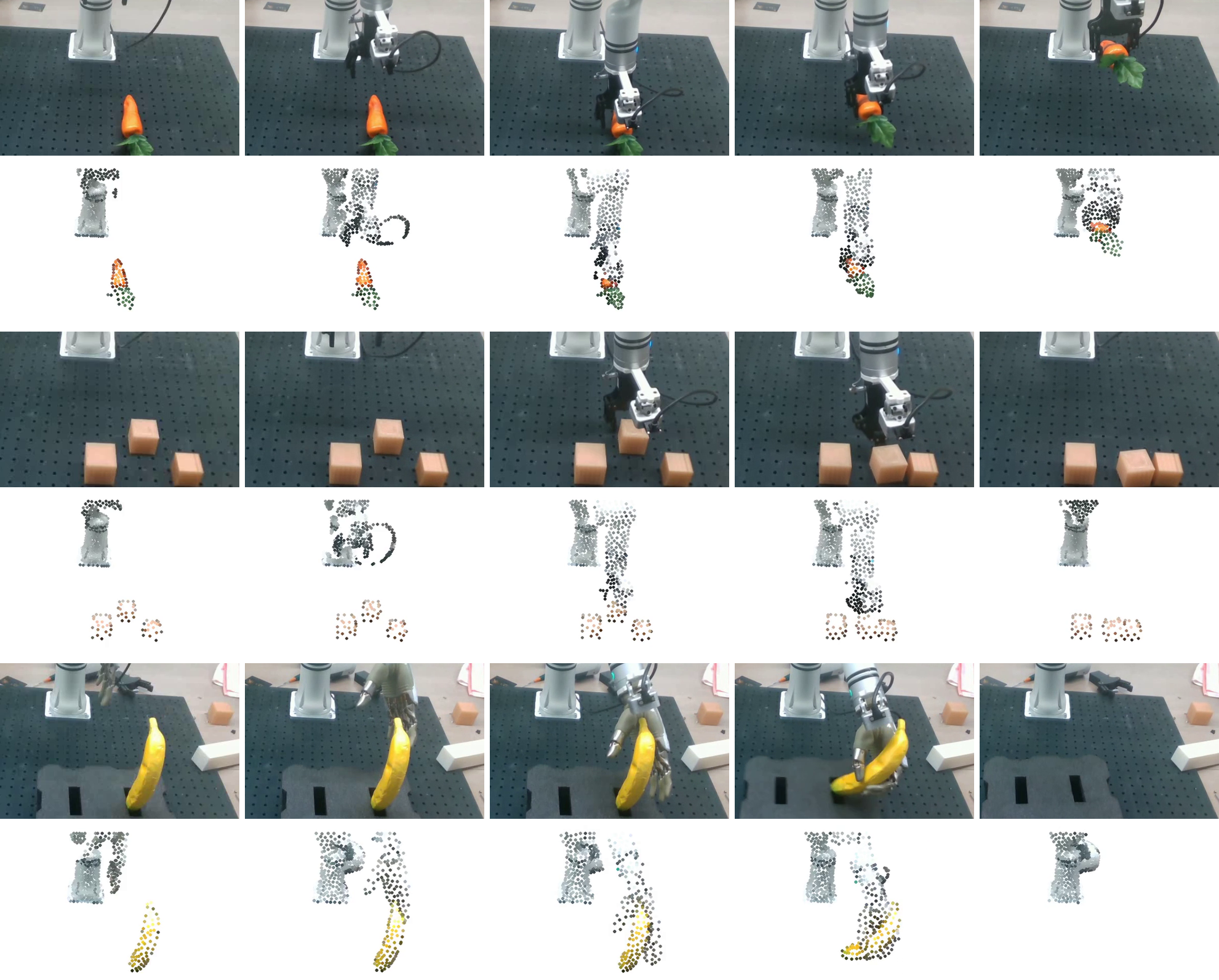}};
  \node[rotate=90] at (0.275,-2.560) {Gripper: pick};
  \node[rotate=90] at (0.275,-7.120) {Gripper: push};
  \node[rotate=90] at (0.275,-11.680) {Hand: pick};
\end{tikzpicture}
\caption{\textbf{Real-robot deployment of JEPA4R on a RealMan RM75-B arm.} Key frames of one execution per task at successive time steps, with the camera view above and the point cloud below: picking and pushing with a parallel gripper, and picking with a dexterous hand.}
\label{fig:real}
\end{figure*}
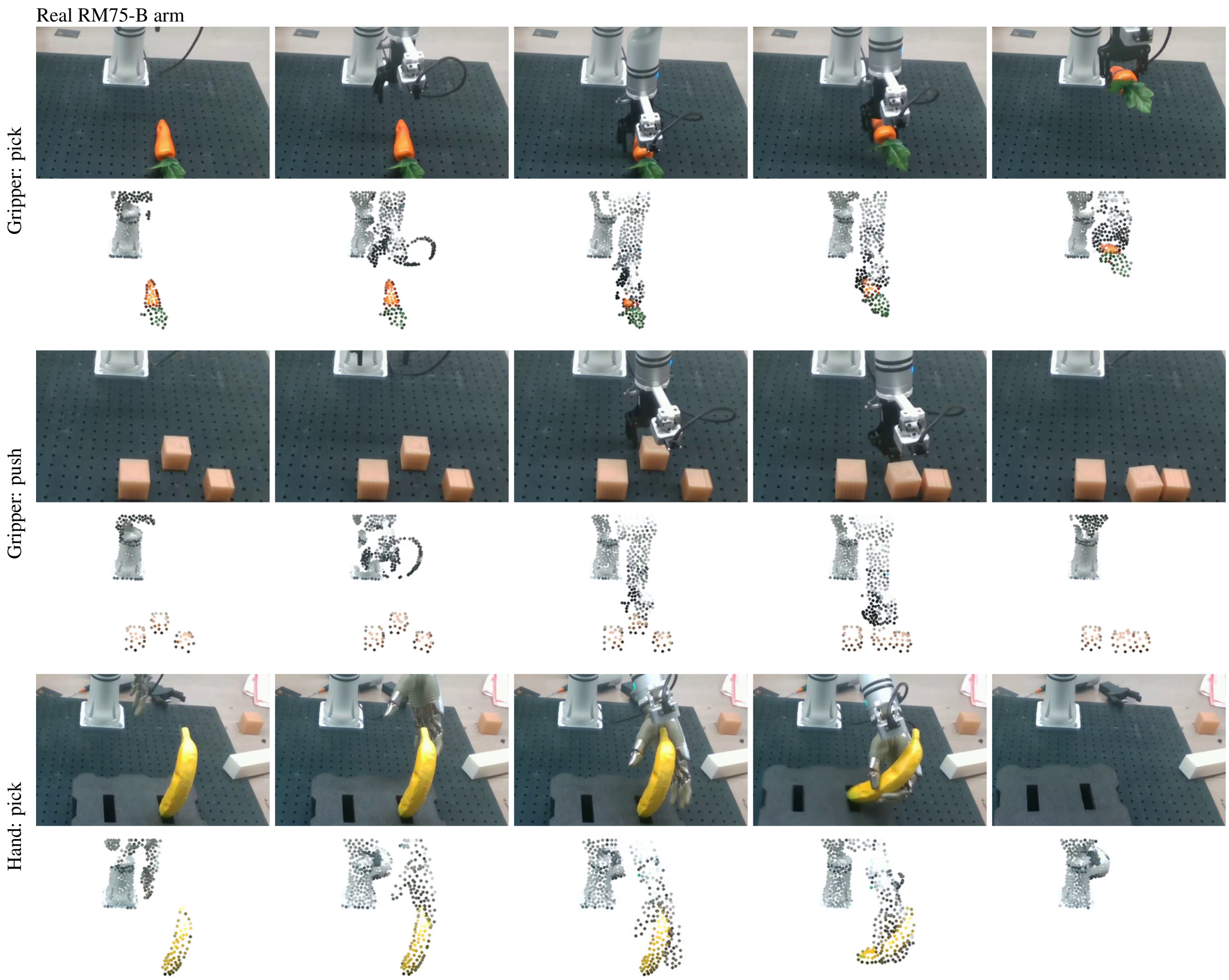

\subsection{Real World Evaluation}
\label{sec:exp-real}

\paragraph{Robot Setup.}
We deploy JEPA4R on a RealMan RM75-B arm to verify that the world model and policy heads trained only in its digital twin run on real hardware. With an Inspire EG2-4C2 parallel gripper, the arm picks up objects and pushes cubes; with an OYMotion AP001 dexterous hand, it pulls objects out of a holder. At every step, JEPA4R encodes the point cloud from a single depth camera, and the policy head outputs the next action.

\paragraph{Results.}
JEPA4R completes all three tasks on the real arm. \cref{fig:real} presents one execution per task, with key frames at successive time steps.

\section{Limitation and Future Work}
\label{sec:discussion}

\looseness=1 This paper focuses on JEPA world models and policy learning from single-camera point clouds. A general world model shared across scenes and robots relies on large-scale, diverse 3D robot data, which remains scarce and is beyond the scope of this paper. How to collect high-quality data for JEPA world models, and whether a general recipe exists, remain open questions and are the focus of our next steps.

\section{Conclusion}
\label{sec:conclusion}

Most JEPA world models have seen scenes in 2D, been judged in simulation by short-horizon latent planning and hidden what they imagine.
JEPA4R predicts, plans and acts in the latent space of single-depth-camera point clouds, and on 3D manipulation tasks it leads the 2D JEPAs by a large margin.
It ranks first on MetaWorld and Adroit Relocate with 98.4\% and 88.0\% success, above the runner-up, R3M, at 96.0\% and 65.3\%.
Its decoder turns every imagined rollout into 4D point clouds, with a mean point error within 1.1\,cm of the observed points over 200 environment steps.
Beyond simulation, JEPA4R completes closed-loop deployment on a real robot arm with both a gripper and a dexterous hand, after training only in a digital twin.
Because manipulation hinges on metric 3D geometry, these results suggest that 3D observations may become the default input of embodied JEPA world models.
We hope JEPA4R brings these models closer to robots that plan in 4D point space and check what they imagine before they act.

\clearpage
\section{LLM Usage Statement}

We used large language models (LLMs) as follows. Claude Code and Codex generated code for the repository. Claude (Anthropic) and GPT (OpenAI) generated drafts of the paper text, including the related-work section, parts of the experimental design, execution and analysis, and figures. We also gave the manuscript to Claude and GPT to obtain reviewer-style feedback. In all cases, the authors reviewed, edited and approved the generated content. No LLMs were used to generate research ideas, hypotheses, claims or conclusions, or to create or annotate data. The authors take full responsibility for all content of this paper.

{
    \small
    \bibliographystyle{ieeenat_fullname}
    \bibliography{refs}
}

\clearpage
\setcounter{page}{1}
\maketitlesupplementary
\setcounter{section}{0}\setcounter{table}{0}\setcounter{figure}{0}\setcounter{equation}{0}
\renewcommand{\thesection}{S\arabic{section}}\renewcommand{\theHsection}{S\arabic{section}}
\renewcommand{\thetable}{S\arabic{table}}\renewcommand{\theHtable}{S\arabic{table}}
\renewcommand{\thefigure}{S\arabic{figure}}\renewcommand{\theHfigure}{S\arabic{figure}}
\renewcommand{\theequation}{S\arabic{equation}}\renewcommand{\theHequation}{S\arabic{equation}}
\makeatletter\renewcommand*{\Hy@backout}[1]{}\makeatother

\section{Planning under LeWM's setting}
\label{sec:supp-planning}

Following LeWM~\cite{maes2026lewm}, we also evaluate both world models by goal-reaching planning. Planning encodes the current observation and a goal $o_g$, unrolls $g_\phi$ over $H=5$ action blocks, and solves
\begin{equation}
  a^{\star}_{t:t+H-1}=\argmin_{a_{t:t+H-1}}\big\lVert\hat z_{t+H}-z_g\big\rVert_2^2,
  \label{eq:cem}
\end{equation}
with goal latent $z_g=f_\theta(o_g)$. The cross-entropy method (CEM)~\cite{deboer2005cem} samples 300 action sequences, refits a Gaussian to the 30 best and repeats 10 times; the robot executes the plan and replans. \Cref{eq:cem} thus selects the actions whose imagined outcome ends closest to the goal. For JEPA4R, a goal point cloud $P_g$ gives $z_g=f_\theta(P_g)$, and CEM minimizes \cref{eq:cem} unchanged, since it acts only on latents.

We run LeWM's goal-reaching planning on OGBench Cube, with the original UR5e and with the RM75-B twin, which replaces the UR5e.
As in LeWM, CEM starts from a dataset state and has 50 steps to reach the observation 25 environment steps later.
Success requires the cube within 4\,cm of its goal (50 starts).
Random actions from the same starts give the random-policy floor.
Scores are means over the 10 epoch checkpoints.

JEPA4R performs on par with LeWM (\cref{tab:planning}): their gaps stay within the variation of 50-start estimates on one task, about $\pm$6 points. On OGBench Cube, our LeWM run, averaged over its checkpoints, is close to the released checkpoint (69.4 \vs 72--74 on our starts).

\section{Experimental settings}
\label{sec:supp-impl}

\subsection{Paired data}
\label{sec:supp-data}
Each world-model dataset is collected in simulation and rendered by one fixed camera into paired images and point clouds that match row by row (\cref{sec:method-pcd}). On Cube, episodes mix low-noise expert, high-noise expert, expert-then-random and random control in proportions 0.2, 0.4, 0.2 and 0.2; Puzzle follows the noisy recipe of OGBench~\cite{park2025ogbench}; on Scene, MetaWorld and Adroit Relocate, they mix noisy expert control, expert control interrupted by random actions, and random actions in proportions 0.5, 0.3 and 0.2. Cube has 10{,}000 episodes per setting, of 200, 500, 1{,}000 and 600 steps for one to four cubes; Puzzle follows OGBench's official scale, with 1{,}000 episodes on $3\times3$ and $4\times4$ and 3{,}000 on $4\times5$, of 1{,}001 steps each; Scene, MetaWorld (per task) and Adroit Relocate have 10{,}000 episodes of 200 steps.

\begin{table}[t]
\centering
\caption{\textbf{World-model planning success under LeWM's setting (\%).} LeWM and JEPA4R share the predictor, objective and CEM planner; the observation and its parameter-matched encoder differ (pixels \vs depth-camera point clouds). OGBench Cube: LeWM protocol (50 starts, goal 25 steps ahead); scores average the 10 epoch checkpoints. Random: a random policy from the same starts. $\Delta$: JEPA4R minus LeWM.}
\label{tab:planning}
\small
\setlength{\tabcolsep}{5pt}
\begin{tabular}{@{}lcccr@{}}
\toprule
Task & Random & LeWM & JEPA4R & $\Delta$ \\
\midrule
\multicolumn{5}{@{}l}{\textit{OGBench Cube}} \\
\quad UR5e & 42.8 & \textbf{69.4} & 65.2 & $-$4.2 \\
\quad RM75-B & 47.0 & 73.8 & \textbf{79.8} & +6.0 \\
\bottomrule
\end{tabular}
\end{table}

\subsection{World models}
\label{sec:supp-wm}
JEPA4R and the LeWM baseline are identical except for the encoder: a PointNet on point clouds for the former and a ViT-tiny/14 on images for the latter. Both are trained with \cref{eq:objective} for 10 epochs (AdamW, learning rate $5{\times}10^{-5}$, batch size 128), one pair per dataset.

\subsection{Policy learning and evaluation}
\label{sec:supp-policy}
Every encoder is frozen, and every benchmark reports success rate. Cube uses GCIQL~\cite{park2025ogbench} and Puzzle uses decoupled Q-chunking (DQC)~\cite{li2025dqc}, both offline goal-conditioned reinforcement learning on OGBench's official play data for $10^6$ steps: besides the latent, the input carries the robot state on Cube and the button lights on Puzzle, the goal gives the target state of the objects or the lights, and evaluation runs 50 episodes on each of the five official tasks, with Puzzle reporting the best of its ten evaluations during training. MetaWorld and Adroit Relocate use the official imitation protocol of CortexBench~\cite{majumdar2023vc1}, and Scene reuses the same code on its five evaluation tasks: behavior cloning for 100 epochs on 25 demonstrations per task (100 on Relocate), scoring the best of 20 evaluations.

\subsection{Baselines}
\label{sec:supp-baselines}
PLDM~\cite{sobal2025pldm} and LeJEPA~\cite{balestriero2025lejepa} are trained on our image data for 10 epochs with their own recipes; VISReg~\cite{wu2026visreg} and V-JEPA~2.1~\cite{murlabadia2026vjepa21} use the authors' released weights and are evaluated by us. All four encoders are frozen, with the same policy learning and evaluation as JEPA4R. PLDM and LeJEPA use a ViT-tiny/14 and VISReg and V-JEPA~2.1 a ViT-B/16; on Cube, PLDM uses an Impala encoder and V-JEPA~2.1 a ViT-L/16.

\providecommand{\kfblock}[4]{%
\begin{tikzpicture}[x=1cm,y=1cm,font=\footnotesize]
\node[anchor=south west,inner sep=0] at (0.45,0.08) {#2};
\node[anchor=north west,inner sep=0] at (0.45,0) {\includegraphics[width=13.8cm]{#1}};
\node[rotate=90] at (0.18,-1.128) {#3};
\node[rotate=90] at (0.18,-3.437) {#4};
\end{tikzpicture}}

\section{Visualizations}
\label{sec:supp-vis}
\begin{figure*}[p]
\centering
\kfblock{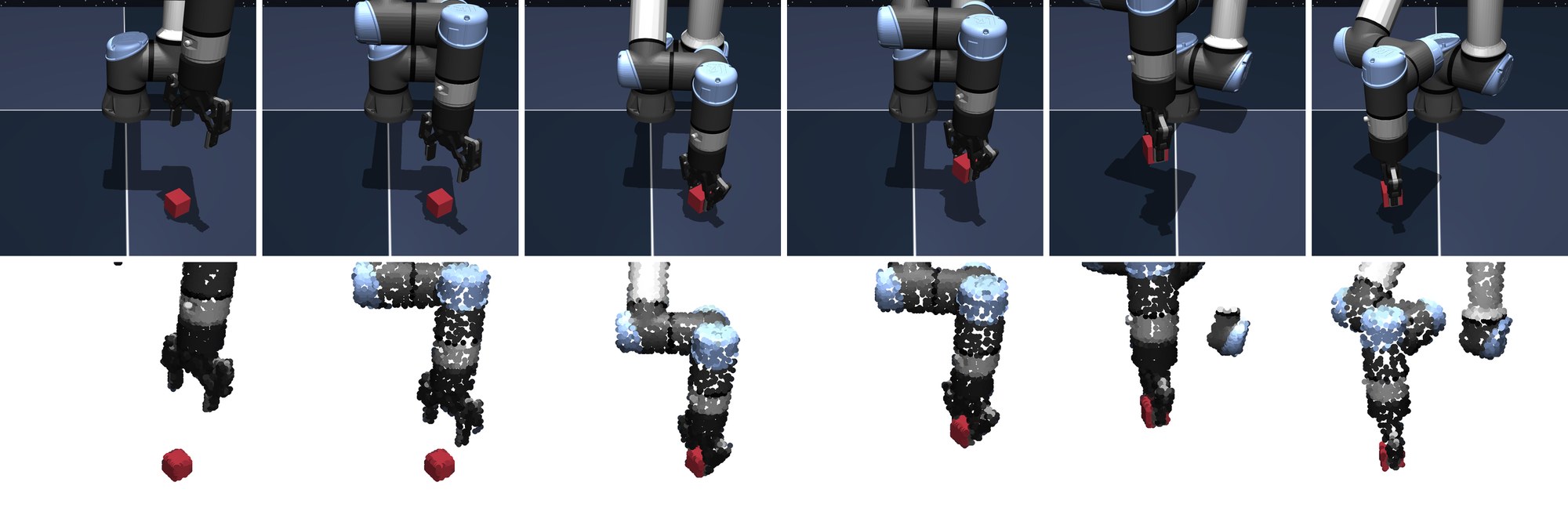}{OGBench Cube-Single}{Camera}{Point cloud}\par\vspace{6pt}
\kfblock{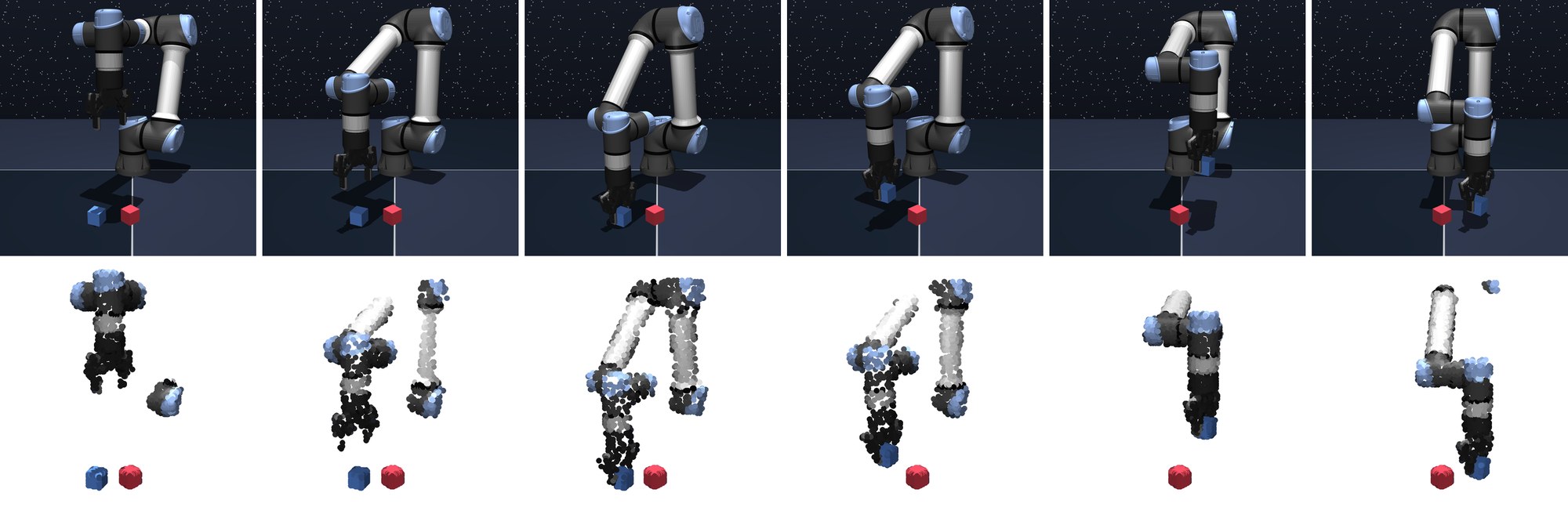}{OGBench Cube-Double}{Camera}{Point cloud}\par\vspace{6pt}
\kfblock{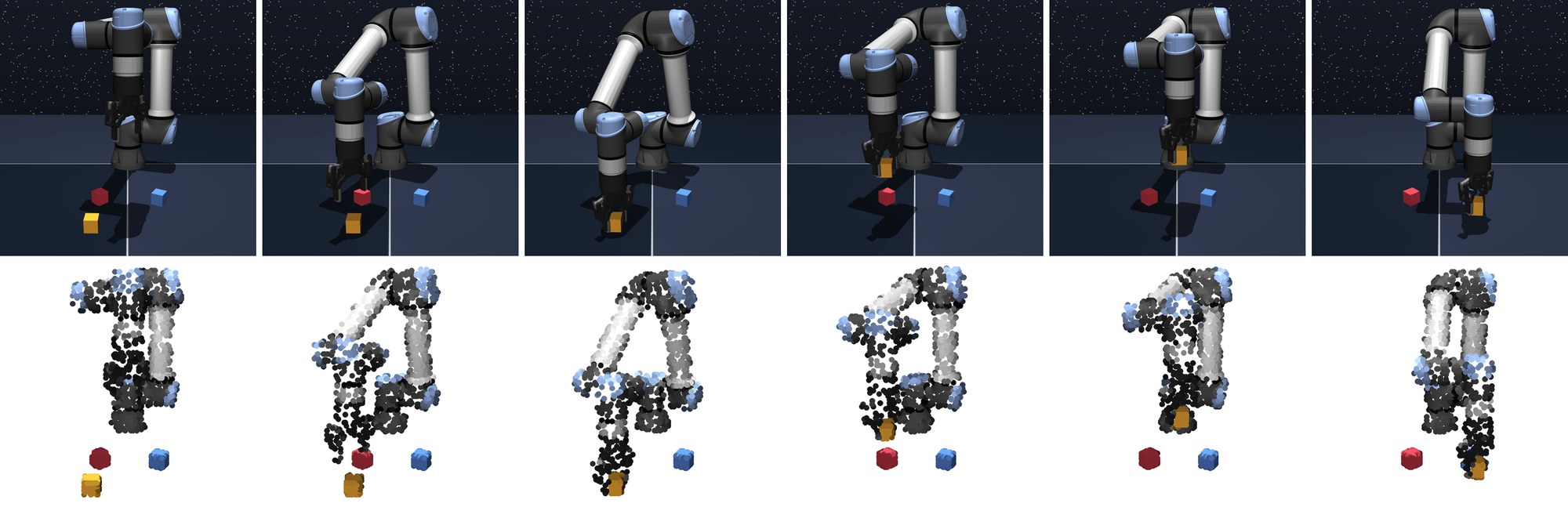}{OGBench Cube-Triple}{Camera}{Point cloud}\par\vspace{6pt}
\kfblock{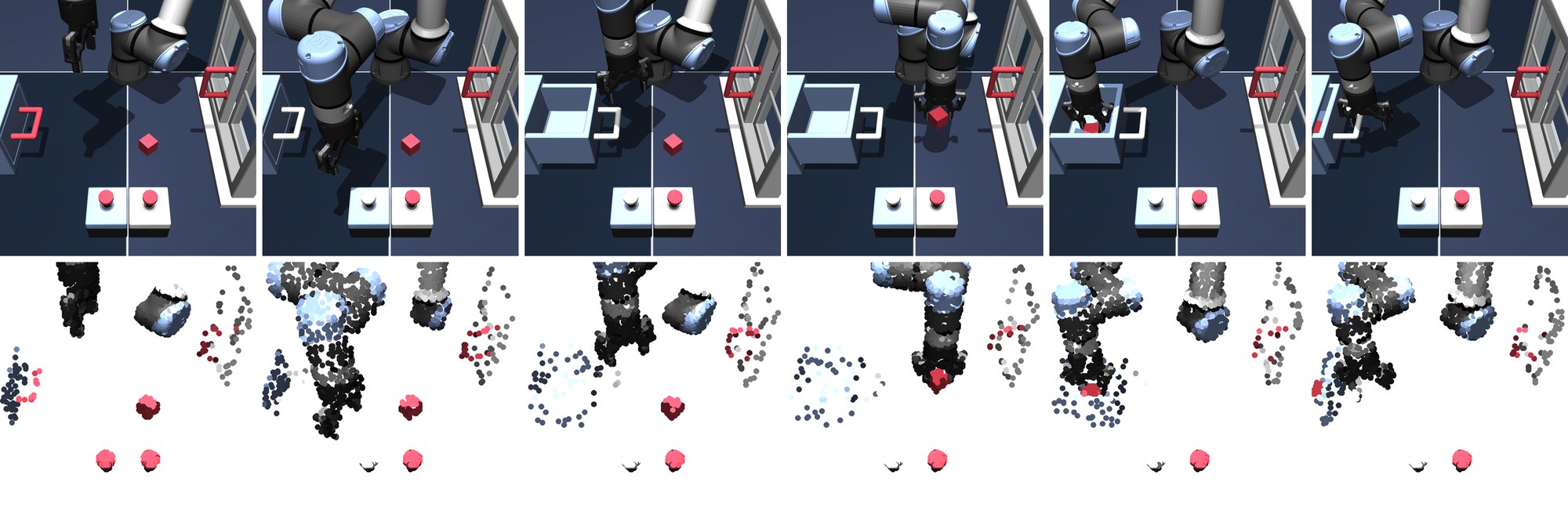}{OGBench Scene}{Camera}{Point cloud}
\caption{\textbf{Key frames on OGBench (I).} Cube-Single, Cube-Double, Cube-Triple and Scene. One successful JEPA4R evaluation episode per task, from the start (left) to the first success (right); top: camera view, bottom: the point cloud JEPA4R receives.}
\label{fig:supp-og1}
\end{figure*}

\begin{figure*}[p]
\centering
\kfblock{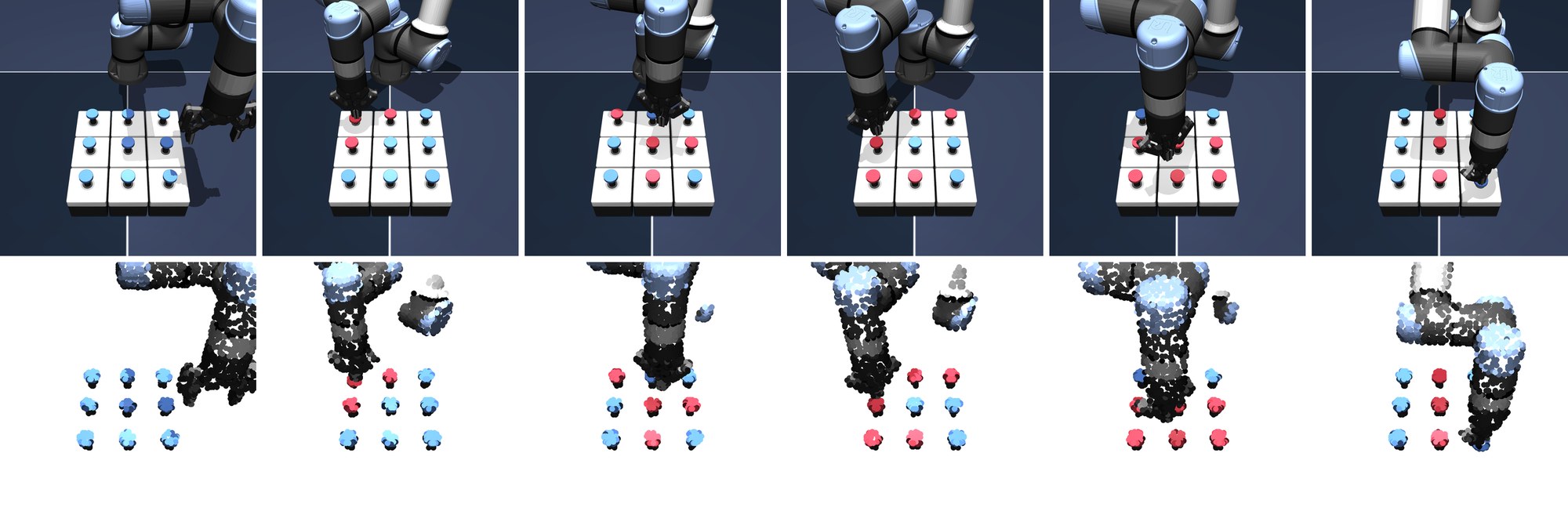}{OGBench Puzzle $3\times3$}{Camera}{Point cloud}\par\vspace{6pt}
\kfblock{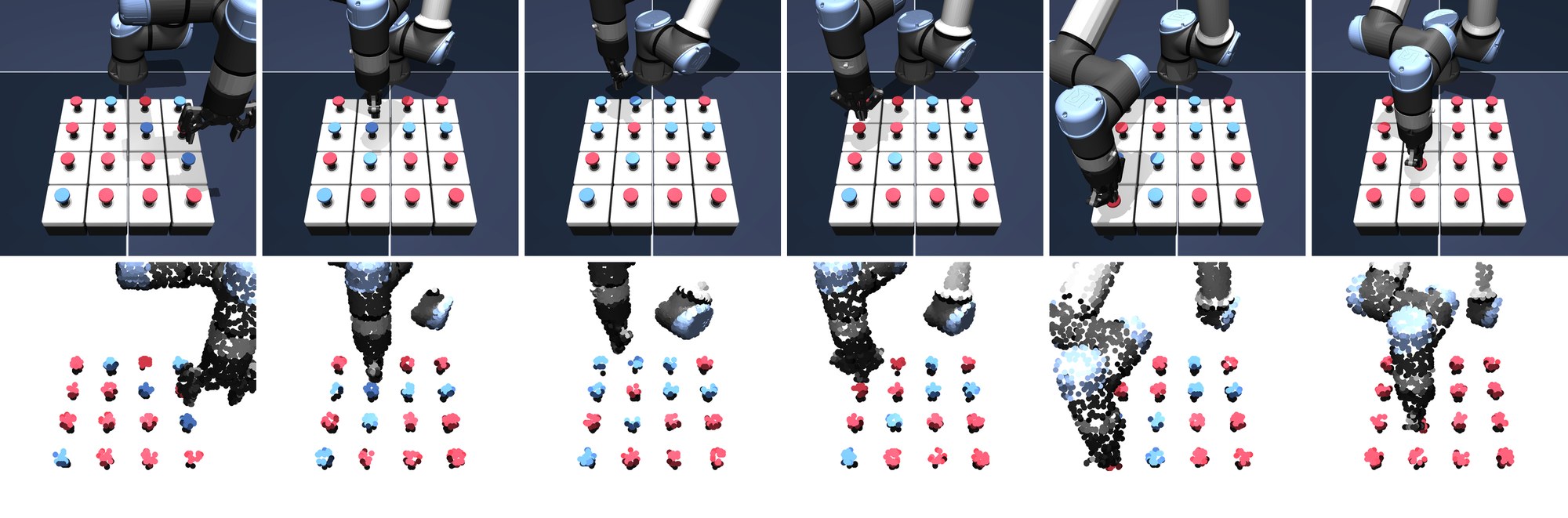}{OGBench Puzzle $4\times4$}{Camera}{Point cloud}\par\vspace{6pt}
\kfblock{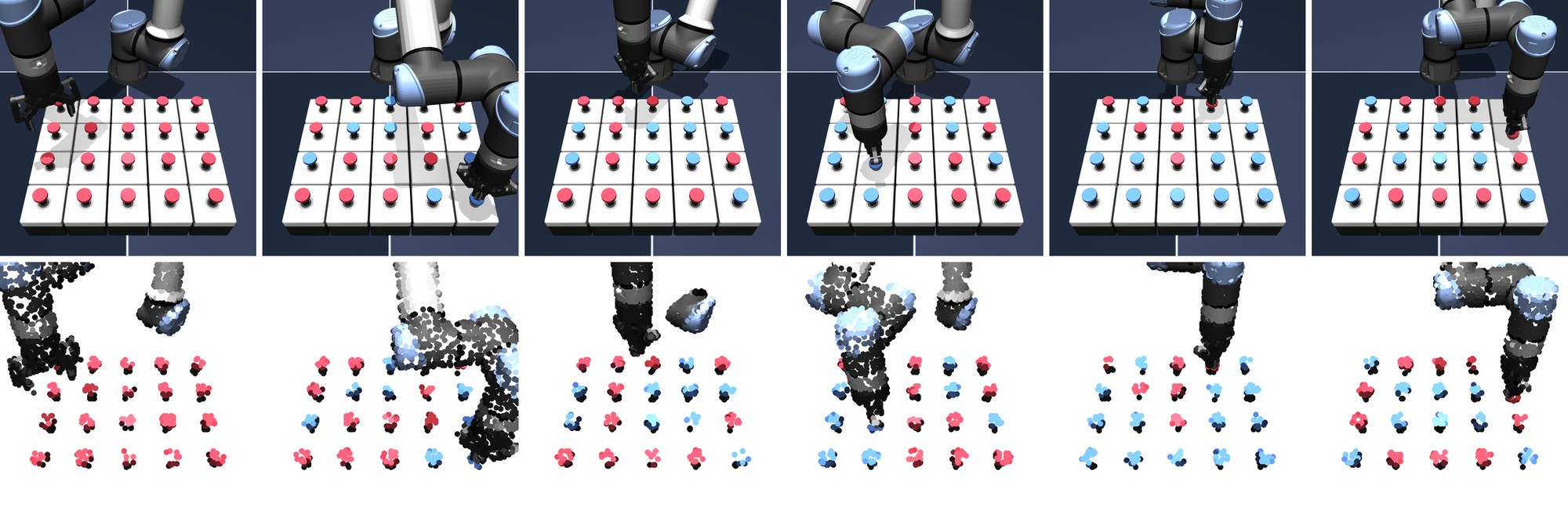}{OGBench Puzzle $4\times5$}{Camera}{Point cloud}
\caption{\textbf{Key frames on OGBench (II).} Puzzle on the $3\times3$, $4\times4$ and $4\times5$ boards, shown as in \cref{fig:supp-og1}.}
\label{fig:supp-og2}
\end{figure*}

\begin{figure*}[p]
\centering
\kfblock{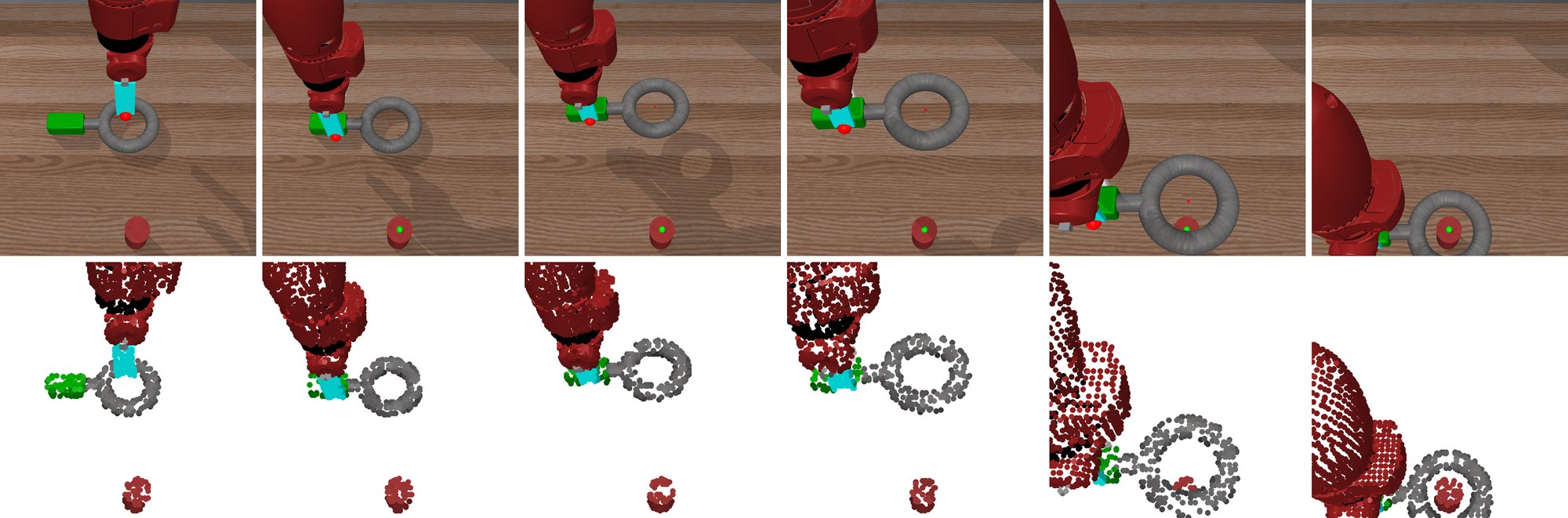}{MetaWorld Assembly}{Camera}{Point cloud}\par\vspace{6pt}
\kfblock{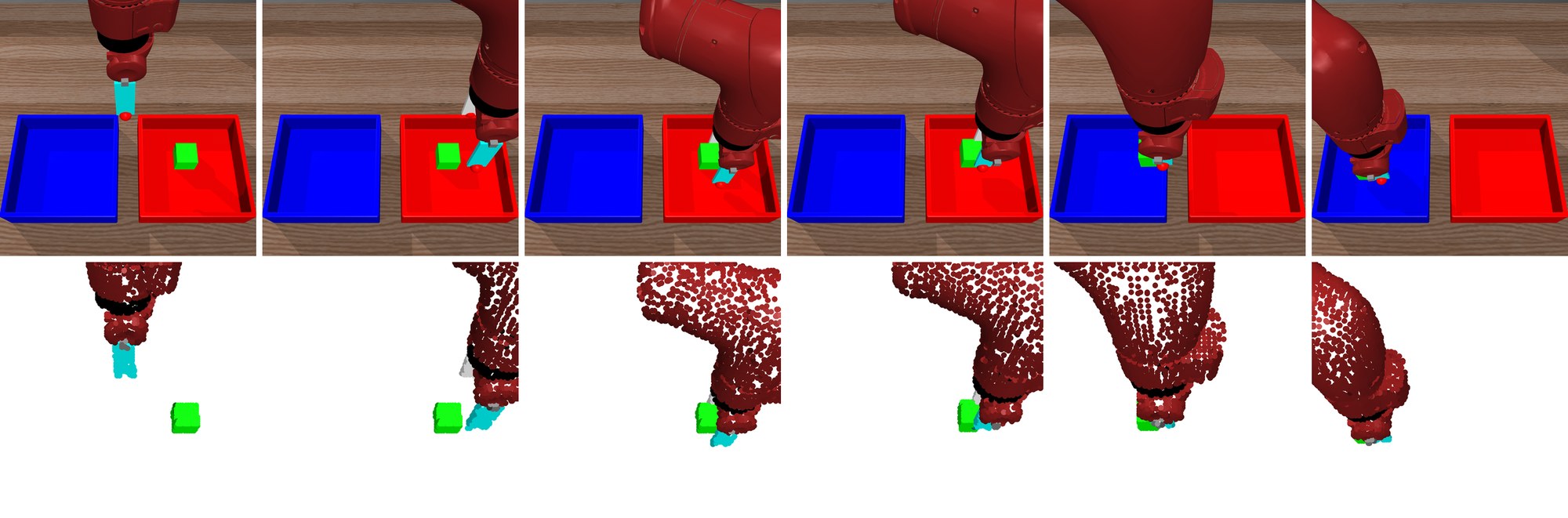}{MetaWorld Bin-pick}{Camera}{Point cloud}\par\vspace{6pt}
\kfblock{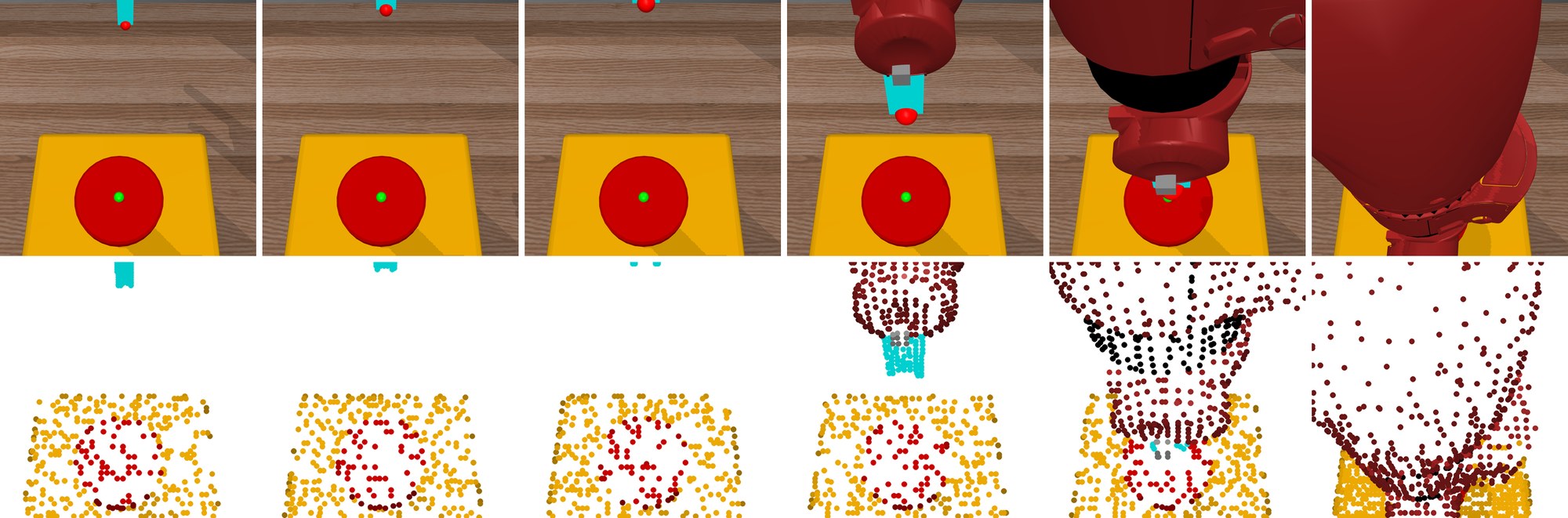}{MetaWorld Button}{Camera}{Point cloud}
\caption{\textbf{Key frames on CortexBench (I).} MetaWorld Assembly, Bin-pick and Button, shown as in \cref{fig:supp-og1}.}
\label{fig:supp-cb1}
\end{figure*}

\begin{figure*}[p]
\centering
\kfblock{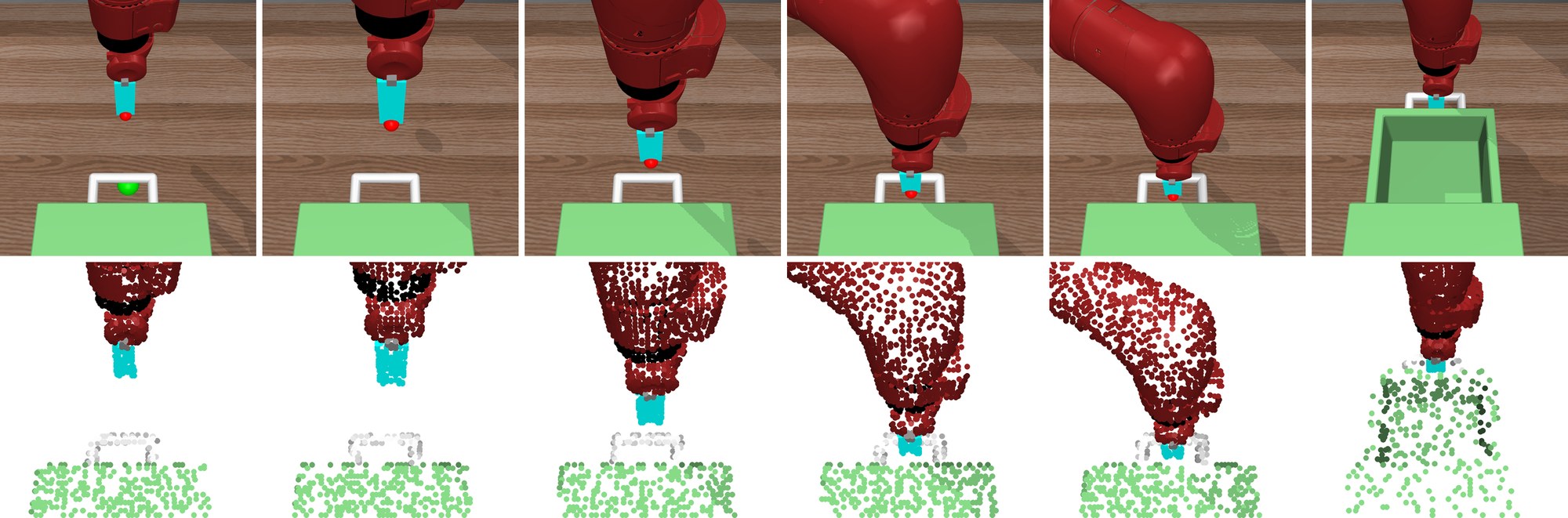}{MetaWorld Drawer}{Camera}{Point cloud}\par\vspace{6pt}
\kfblock{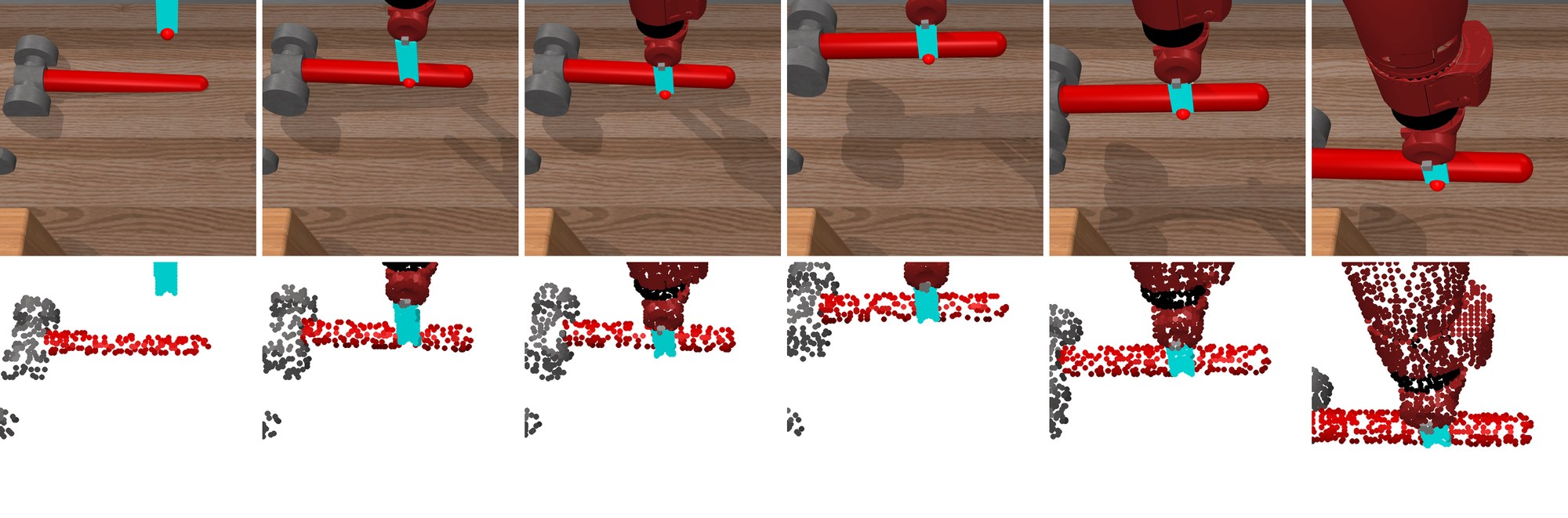}{MetaWorld Hammer}{Camera}{Point cloud}\par\vspace{6pt}
\kfblock{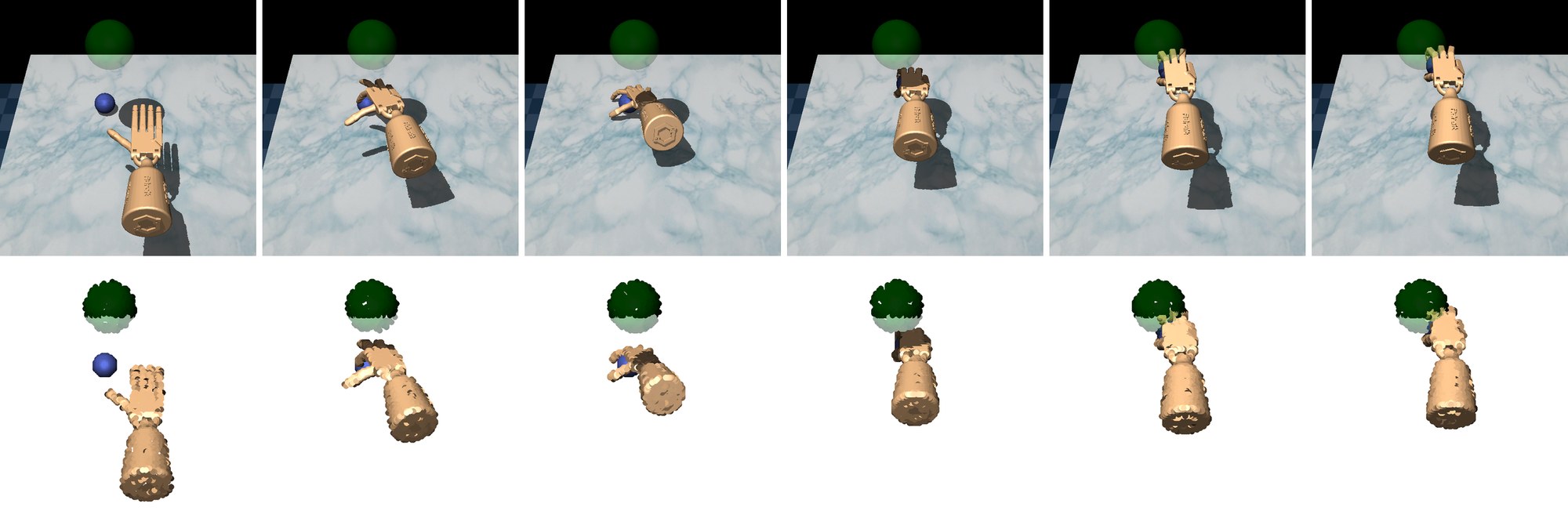}{Adroit Relocate}{Camera}{Point cloud}
\caption{\textbf{Key frames on CortexBench (II).} MetaWorld Drawer and Hammer, and Adroit Relocate, shown as in \cref{fig:supp-og1}.}
\label{fig:supp-cb2}
\end{figure*}

\begin{figure*}[p]
\centering
\kfblock{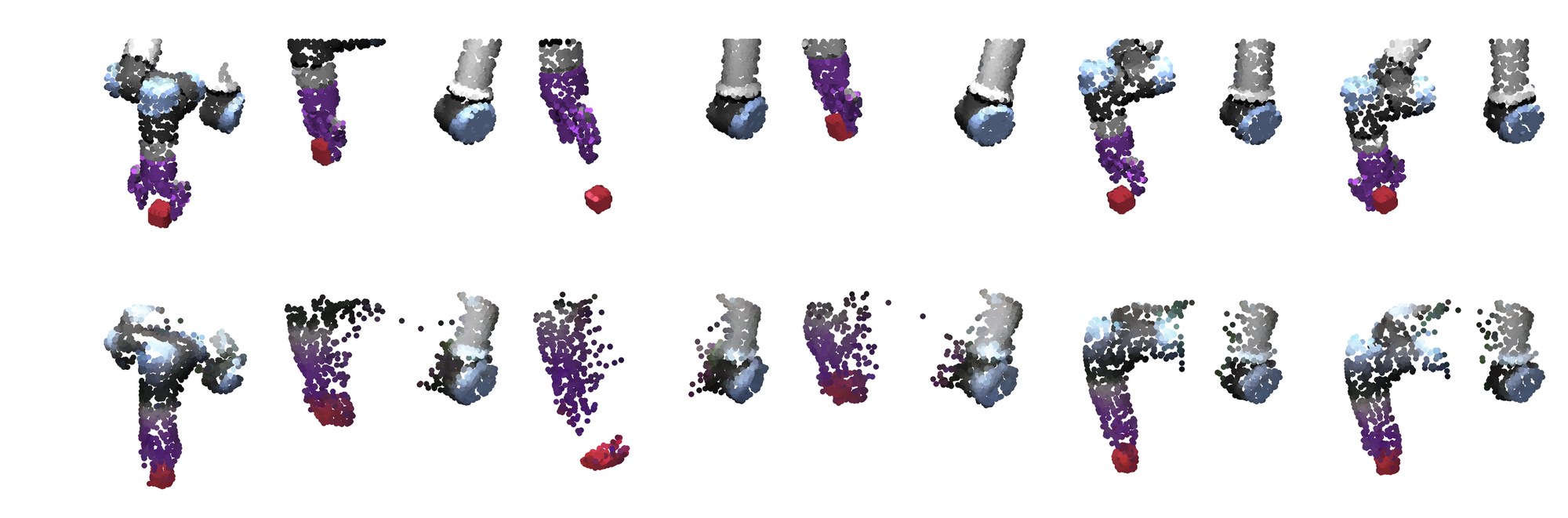}{Example 1}{Ground truth}{Imagined}\par\vspace{6pt}
\kfblock{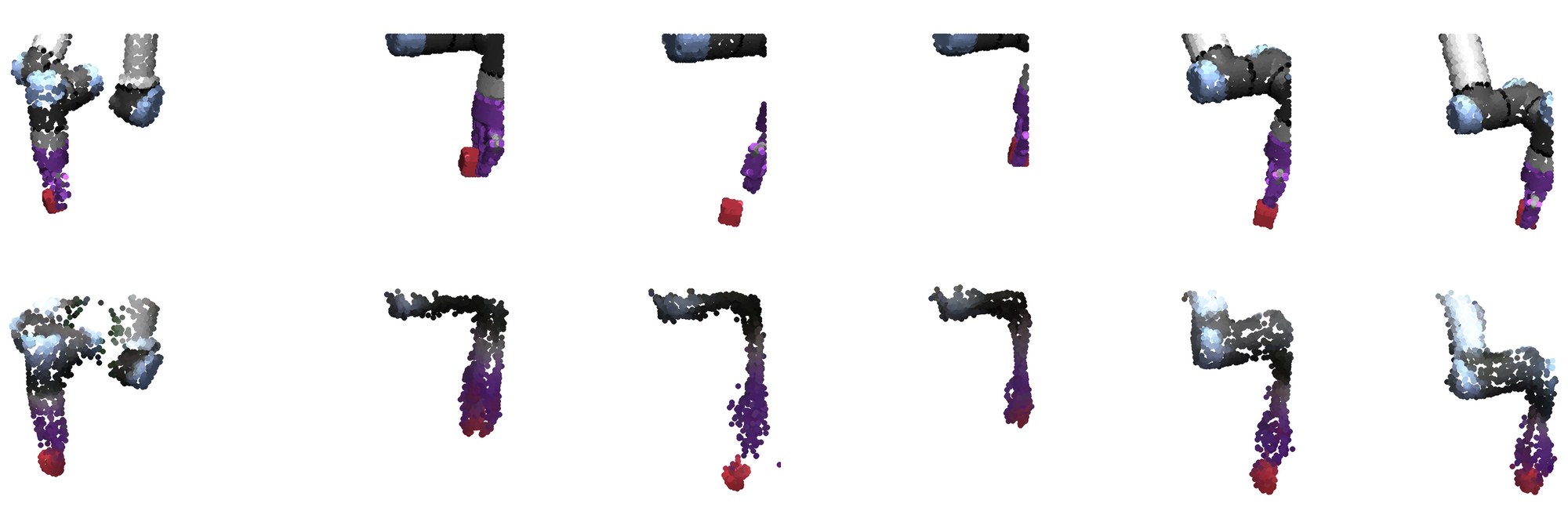}{Example 2}{Ground truth}{Imagined}\par\vspace{6pt}
\kfblock{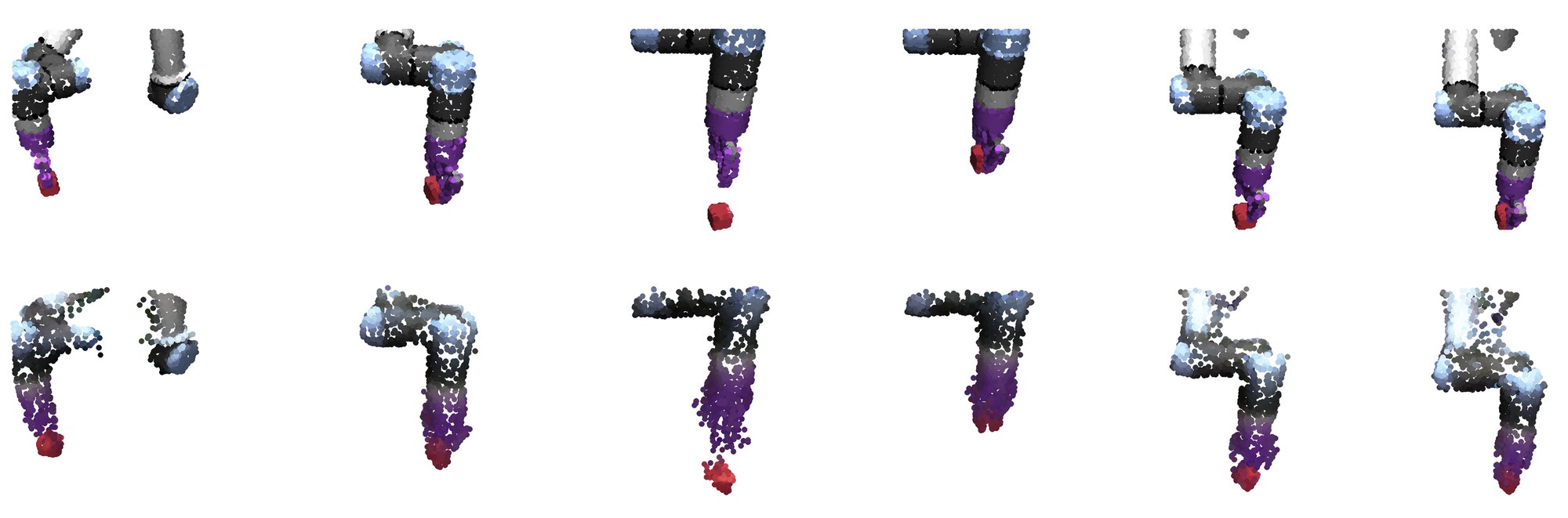}{Example 3}{Ground truth}{Imagined}
\caption{\textbf{Decoded imagination on OGBench Cube-Single (I).} Test episodes. The world model receives three context frames and then predicts from the recorded actions alone; top: ground-truth point clouds, bottom: point clouds decoded from the imagined latents, at environment steps 20 to 200.}
\label{fig:supp-dec1}
\end{figure*}

\begin{figure*}[p]
\centering
\kfblock{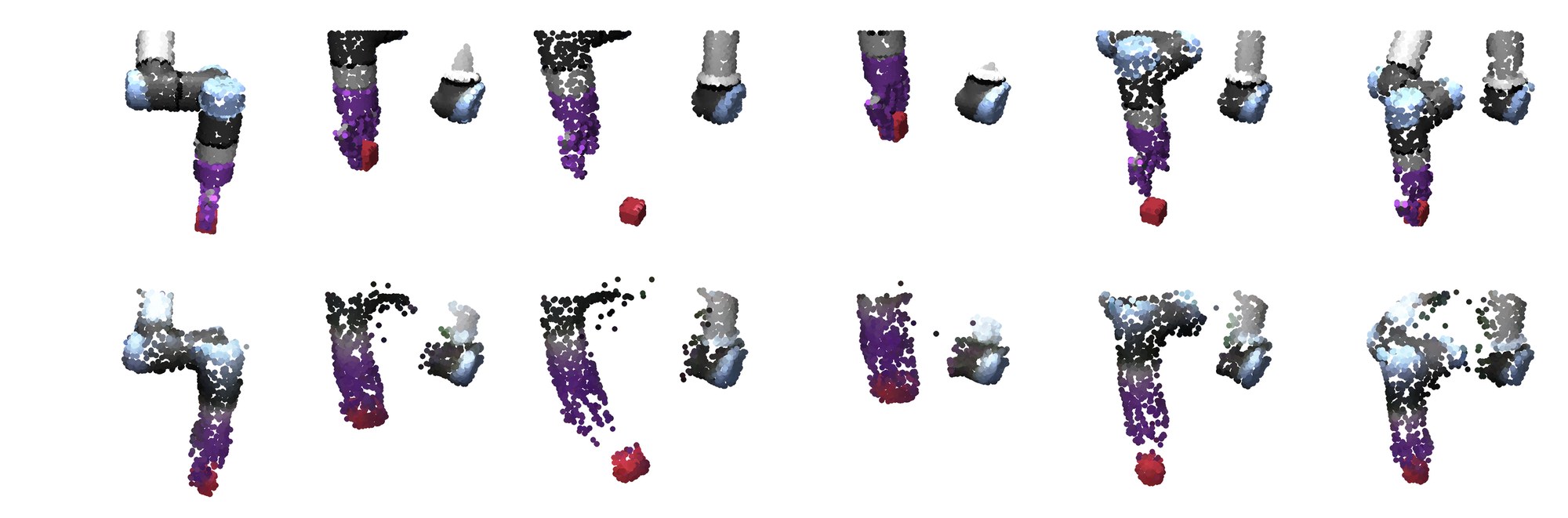}{Example 4}{Ground truth}{Imagined}\par\vspace{6pt}
\kfblock{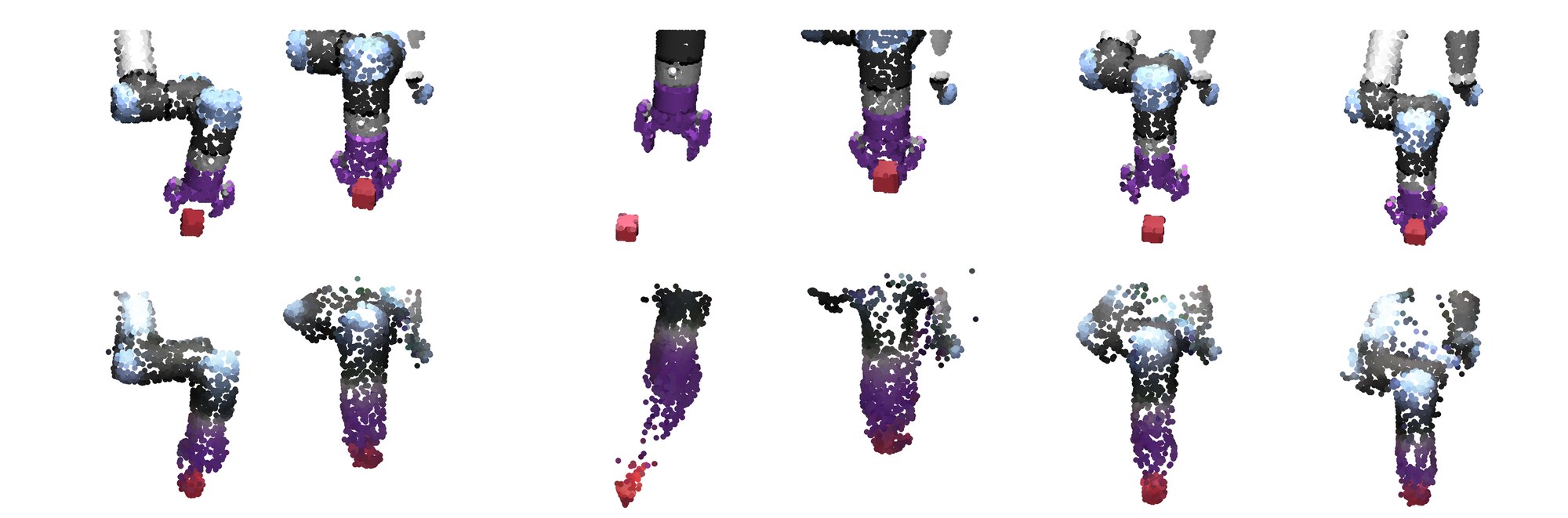}{Example 5}{Ground truth}{Imagined}\par\vspace{6pt}
\kfblock{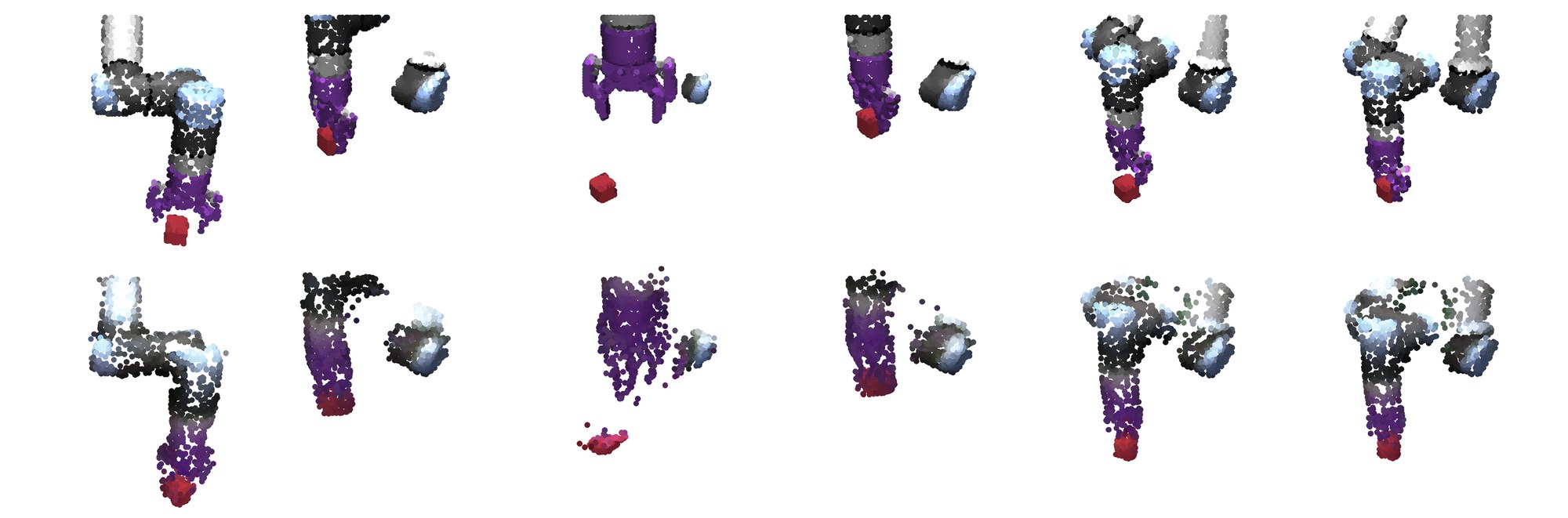}{Example 6}{Ground truth}{Imagined}
\caption{\textbf{Decoded imagination on OGBench Cube-Single (II).} Three more test episodes, shown as in \cref{fig:supp-dec1}.}
\label{fig:supp-dec2}
\end{figure*}

\begin{figure*}[p]
\centering
\kfblock{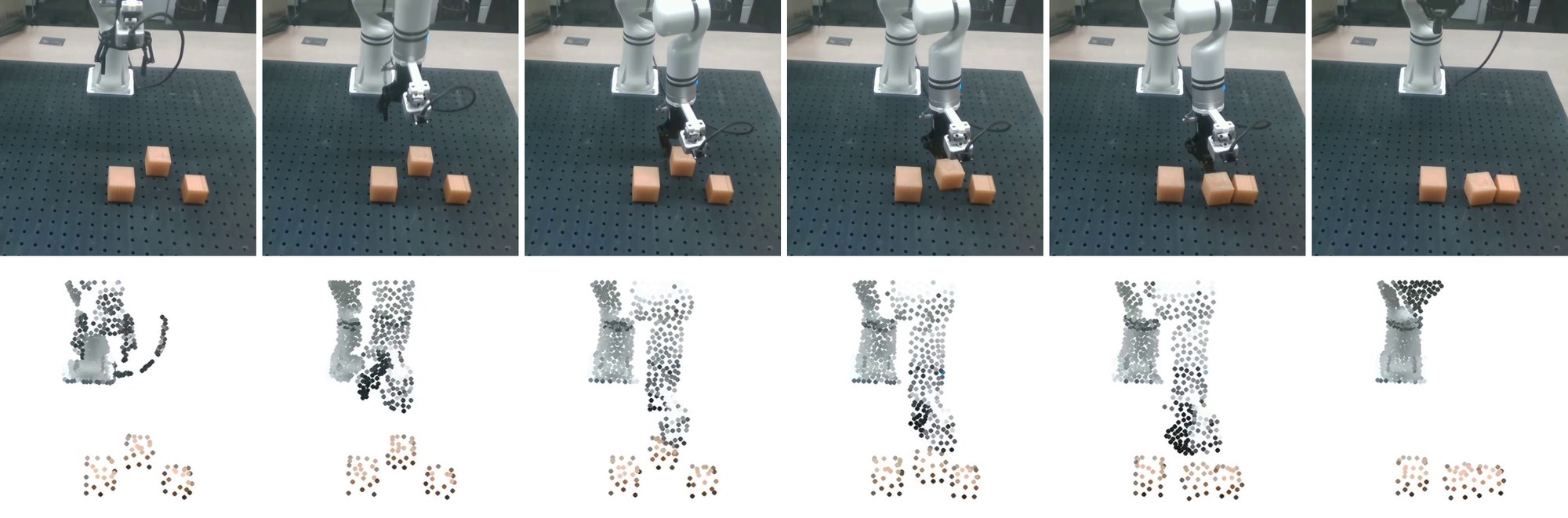}{Parallel gripper: push cubes}{Camera}{Point cloud}\par\vspace{6pt}
\kfblock{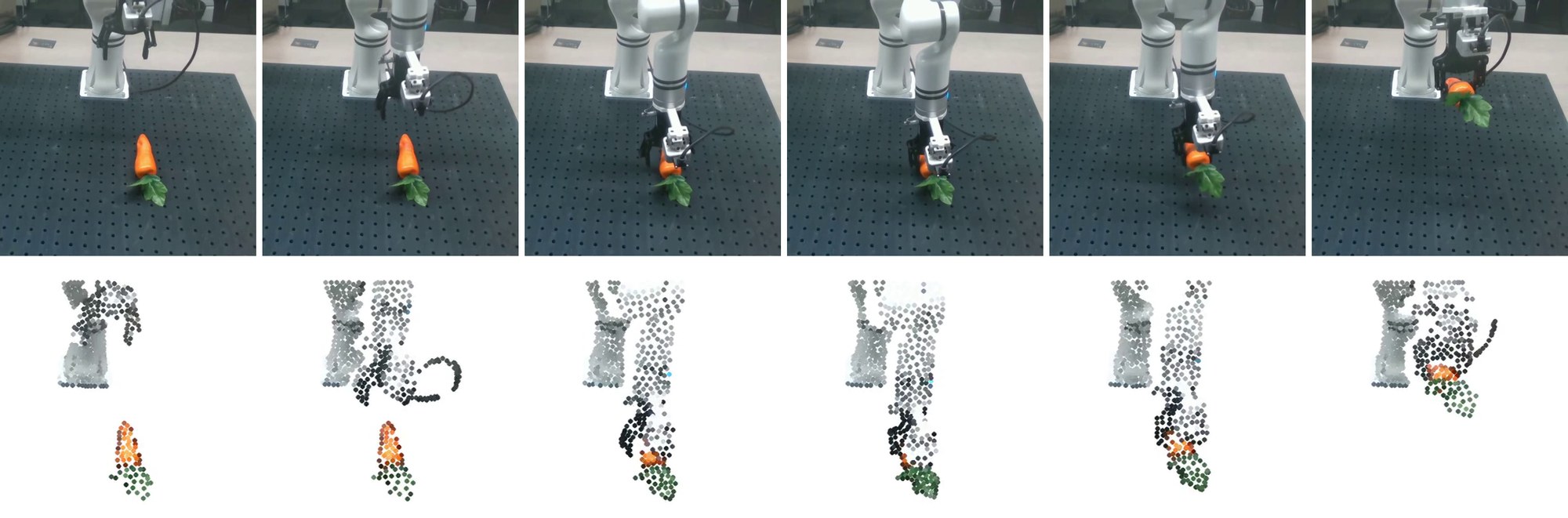}{Parallel gripper: pick a carrot}{Camera}{Point cloud}\par\vspace{6pt}
\kfblock{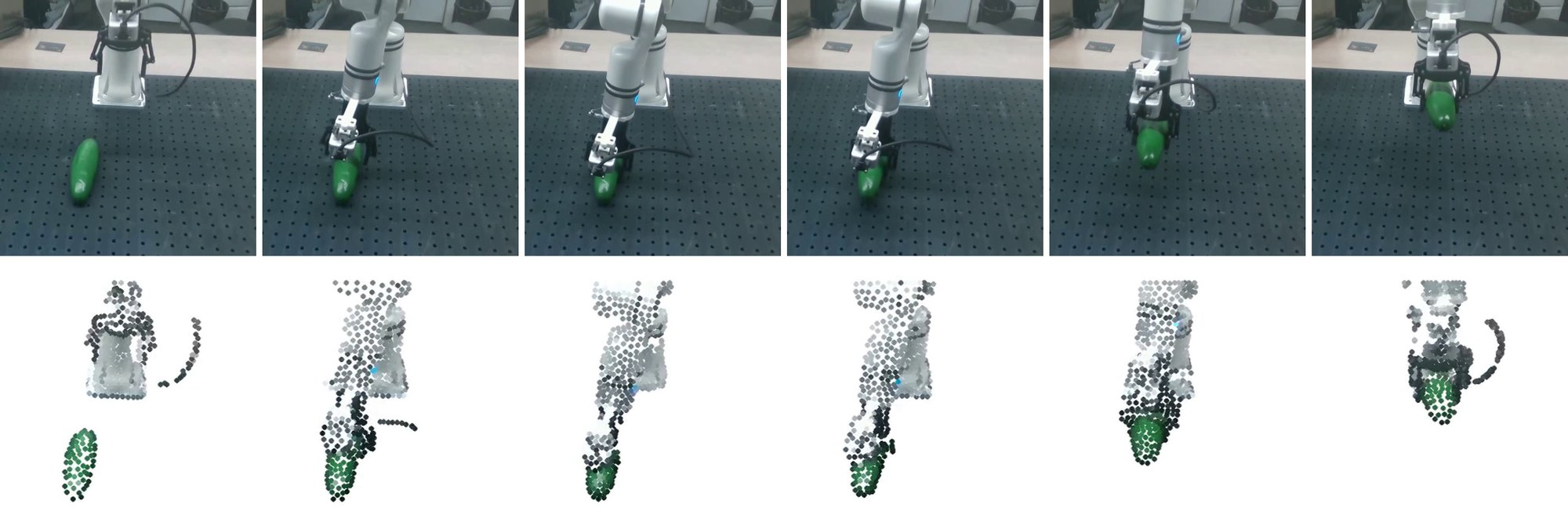}{Parallel gripper: pick a cucumber}{Camera}{Point cloud}\par\vspace{6pt}
\kfblock{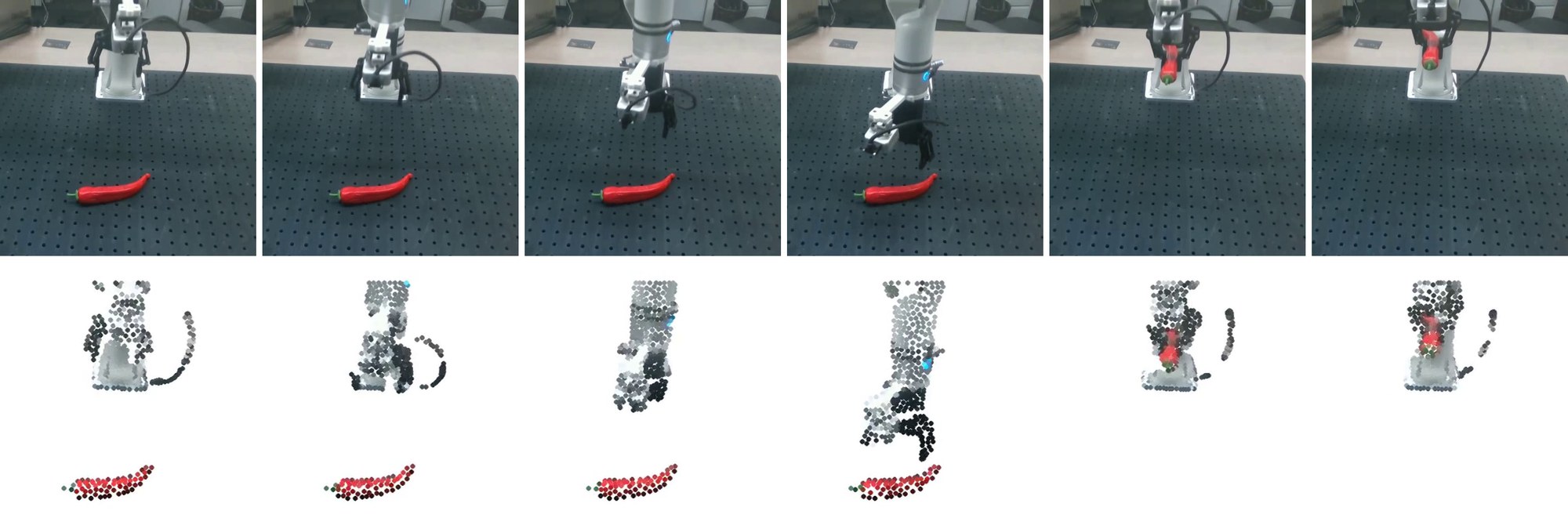}{Parallel gripper: pick a chili}{Camera}{Point cloud}
\caption{\textbf{Real-robot executions with the parallel gripper.} Pushing cubes together and picking up a carrot, a cucumber and a chili on the RealMan RM75-B arm; top: camera view, bottom: point cloud.}
\label{fig:supp-real1}
\end{figure*}

\begin{figure*}[p]
\centering
\kfblock{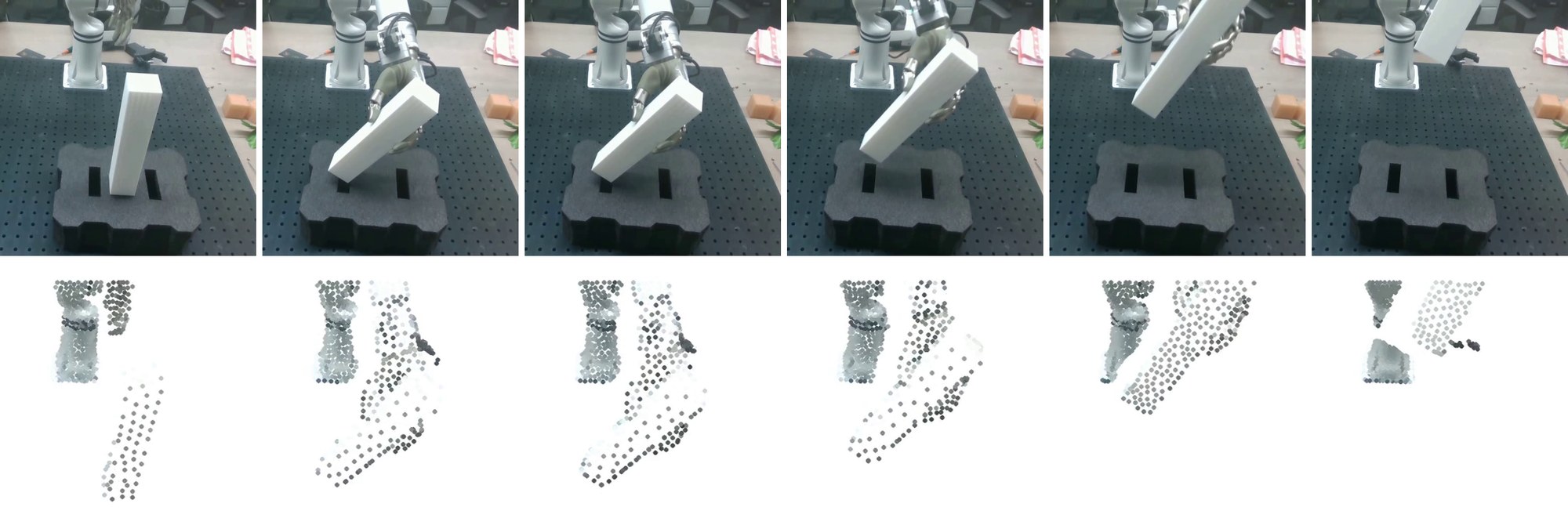}{Dexterous hand: pull a bar}{Camera}{Point cloud}\par\vspace{6pt}
\kfblock{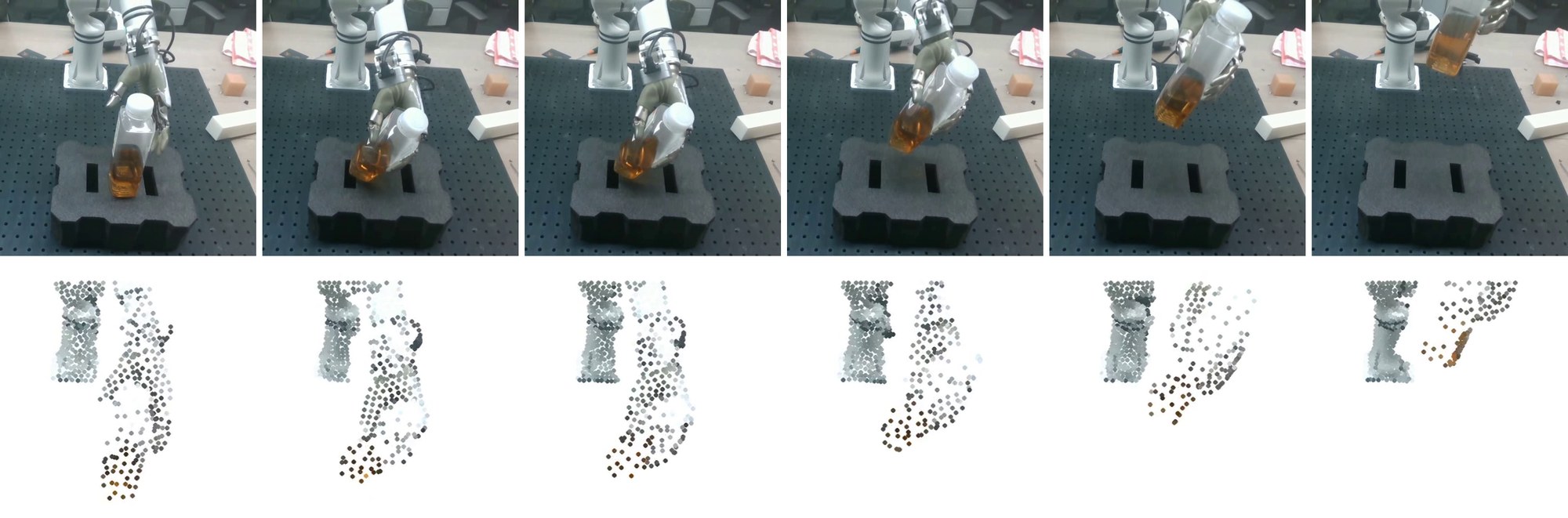}{Dexterous hand: pull a bottle}{Camera}{Point cloud}\par\vspace{6pt}
\kfblock{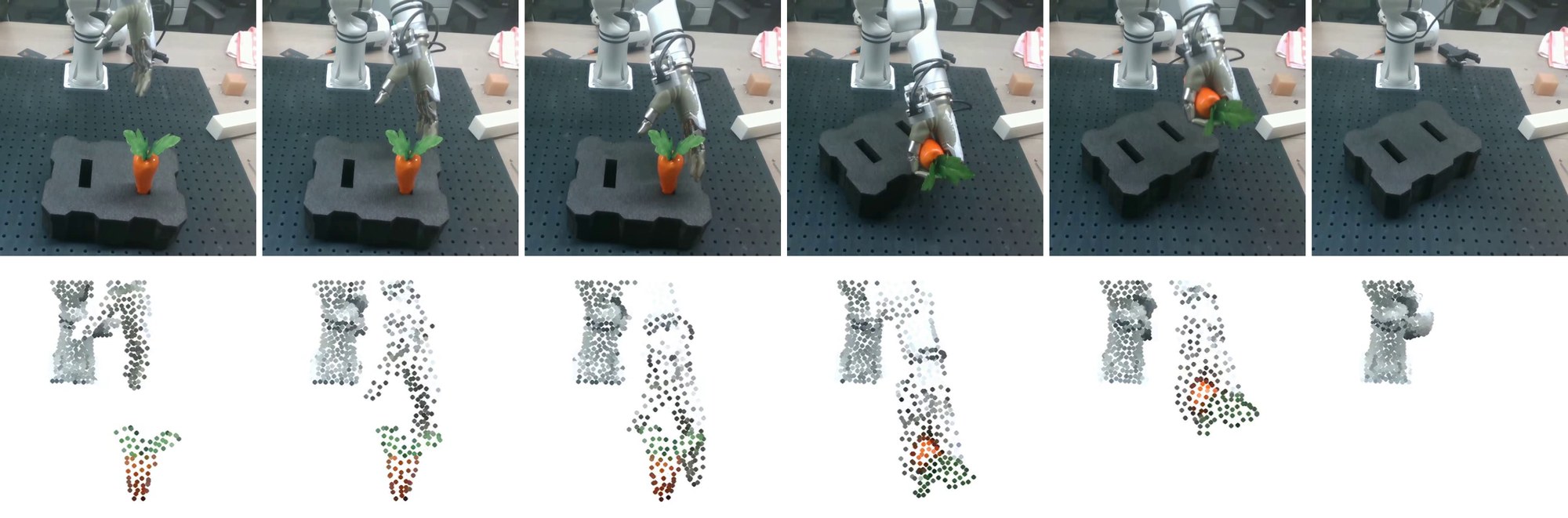}{Dexterous hand: pull a carrot}{Camera}{Point cloud}\par\vspace{6pt}
\kfblock{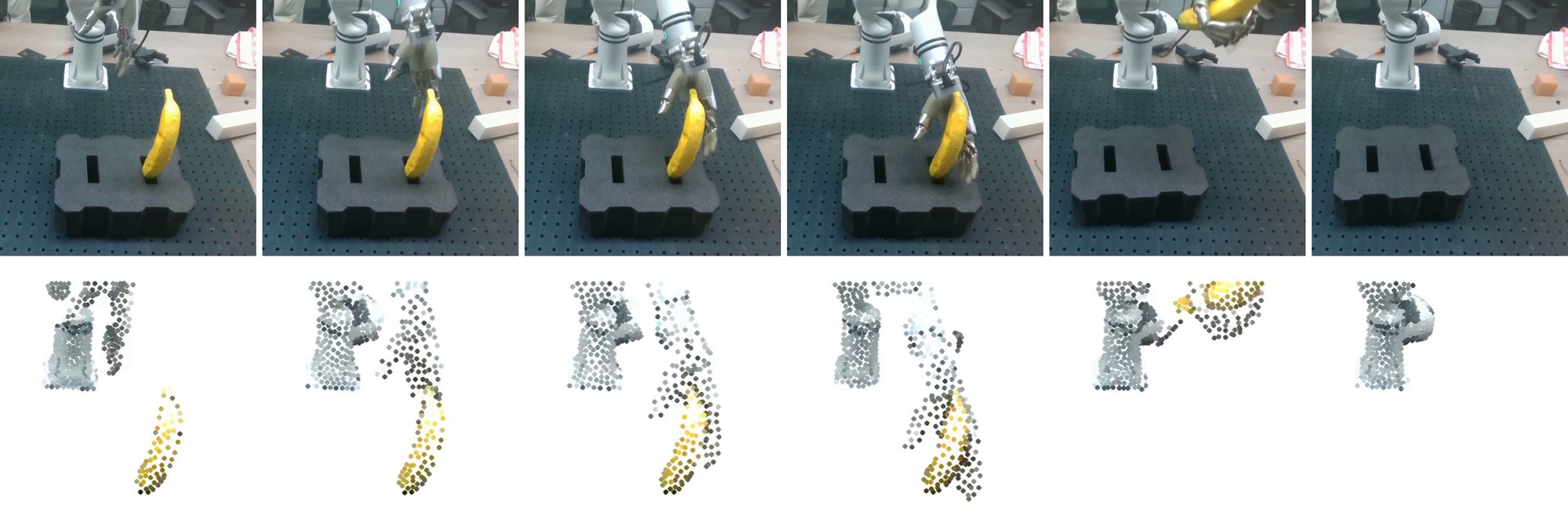}{Dexterous hand: pull a banana}{Camera}{Point cloud}
\caption{\textbf{Real-robot executions with the dexterous hand.} Pulling a bar, a bottle, a carrot and a banana out of a slotted holder, shown as in \cref{fig:supp-real1}.}
\label{fig:supp-real2}
\end{figure*}

We show one successful episode of JEPA4R for every simulated task (\cref{fig:supp-og1,fig:supp-og2,fig:supp-cb1,fig:supp-cb2}), more decoded imagination on OGBench Cube-Single (\cref{fig:supp-dec1,fig:supp-dec2}) and all eight real-robot executions (\cref{fig:supp-real1,fig:supp-real2}). In every figure, frames run from left to right; the camera view is shown for reference, and JEPA4R receives the point cloud below it.

\paragraph{Simulated tasks.}
On OGBench, a UR5e arm picks and places one, two or three cubes into a goal arrangement on Cube; on the Scene task shown, it presses a button to unlock a drawer, opens the drawer, puts the cube inside and closes it; on Puzzle, it presses buttons that toggle the neighboring lights until the board matches the goal pattern. On MetaWorld, a Sawyer arm puts a ring onto a peg (Assembly), moves a cube from one bin to the other (Bin-pick), presses a button from the top (Button), opens a drawer (Drawer) and drives a nail with a hammer (Hammer). On Adroit Relocate, a dexterous hand picks up a ball and moves it to a target. Each episode runs from the start to the first success.

\paragraph{Decoded imagination.}
As in \cref{fig:recon}, the world model receives three context frames and then predicts from the recorded actions alone, and the decoder turns every imagined latent into a point cloud. In each of the six test episodes, the cube starts on the table, is lifted and is put down again.

\paragraph{Real robot.}
All eight executions run zero-shot on the RealMan RM75-B arm with policies trained only in its digital twin. With the Inspire EG2-4C2 parallel gripper, the arm pushes cubes together and picks up a carrot, a cucumber and a chili; with the OYMotion AP001 dexterous hand, it pulls a bar, a bottle, a carrot and a banana out of a slotted holder.

\section{Discussion}
\paragraph{What the results mean.}
The results argue for three changes to the JEPA route. First, 3D-native input: an image entangles metric geometry with appearance and perspective, so a pixel model must infer depth, whereas a point cloud records visible surfaces at their metric positions; the decoding results in \cref{sec:exp-recon} also show that frozen point-cloud latents retain this metric geometry, from which the policy head can read where the objects and the gripper are in space. Second, evaluation through the policies robots learn, rather than through short-horizon goal reaching alone: latents that plan well can transfer poorly to policy learning~\cite{liu2026tcsigreg}, so evaluating by planning alone does not show whether a representation serves policy learning. Third, decoded latents as a standard diagnostic: frozen latents decode back to point clouds, so imagined futures can be compared directly with observations.

\paragraph{Outlook.}
The approach is promising in several directions. On the 3D representation that JEPA4R learns, a policy can be trained from little data and deployed on the real arm, which suits robotic settings where real data are expensive and hard to collect at scale; the world model and the policy head, trained only in a digital twin, run zero-shot on the real arm, so embodied world models can be trained mainly in digital twins rather than on large amounts of real robot video. The JEPA recipe carries over to point clouds with the predictor and the objective unchanged, the same upgrade applies to other JEPA world models, and 3D observations may become the default input of embodied JEPA world models. The decoder shows that, while training stays free of reconstruction, frozen latents and entire imagined rollouts can still be decoded into metrically accurate 4D point clouds, so a robot can check the expected outcome of an action sequence before executing it; building on this, planning can move from approaching a goal latent to operating in 4D point space, with goals given directly as point clouds and every imagined step checkable in metric space. Finally, combining paired 2D/3D data with policy learning offers a way to compare world-model representations beyond short-horizon planning.

\end{document}